\documentclass{eitnlp}

\title{Perceive, Solve, Reason: Progressive Capability Formation in Time Series-LLMs}
\runningtitle{Perceive, Solve, Reason: Progressive Capability Formation in Time Series-LLMs}
\author{\textbf{Xin Qiu}\textsuperscript{1,2}\qquad
        \textbf{Junlong Tong}\textsuperscript{1,3}\qquad
        \textbf{Wei Zhang}\textsuperscript{1}\qquad
        \textbf{Xiaoyu Shen}\textsuperscript{1\textdagger}}
\affiliation{%
  \textsuperscript{1}EIT-NLP Lab, Eastern Institute of Technology, Ningbo\\
  \textsuperscript{2}Zhejiang University\qquad
  \textsuperscript{3}Shanghai Jiao Tong University}
\authornotes{\textsuperscript{\textdagger}Corresponding author.}
\contact{\texttt{qiuxin668@gmail.com,\quad xyshen@eitech.edu.cn}}
\usepackage{hyperref}
\usepackage{url}
\usepackage{graphicx}
\usepackage{booktabs} 
\usepackage{multirow} 
\usepackage{wrapfig}
\usepackage{caption}
\usepackage{float}
\usepackage{makecell}
\usepackage[most]{tcolorbox}
\definecolor{green_negative}{HTML}{37B24D}
\definecolor{mygreen}{HTML}{07ad25}
\definecolor{myred}{HTML}{ee1f1f}

\definecolor{SBCheckGreen}{HTML}{2E8B57}
\definecolor{SBDashGray}{HTML}{A7ACB5}
\definecolor{lightyellow}{RGB}{255, 244, 230}
\definecolor{mycyan}{HTML}{EBFBEE}
\definecolor{mycyan1}{HTML}{2F9E44}

\definecolor{mycyan3}{HTML}{FFF0F6}
\definecolor{mycyan4}{HTML}{C92A2A}

\definecolor{mycyan5}{HTML}{F1F3F5}
\definecolor{mycyan6}{HTML}{868E96}
\definecolor{answergreen}{HTML}{69DB7C}

\begin{document}
\maketitle
\begin{abstract}
Large language models (LLMs) can now perceive, describe, and reason about images and audio, yet time series understanding remains underestimated: most work uses LLMs as forecasters and classifiers, reasoning studies are few and narrow, and no suite trains and evaluates the full range of capabilities. In this paper, we build the first such suite, with 337K training and 44K evaluation samples across three progressive capabilities, Perceive, Solve, and Reason, and use it to ask how time series should enter an LLM that must master all three, and in what order. We find that a dedicated encoder is essential: with one, a 4B model surpasses GPT-4o on every perception attribute we test, whereas textualizing the series costs 8--10 points in reasoning accuracy. The capabilities then build on one another: learning them in order beats mixing the same data and rivals reasoning-only training with over three times the reasoning data, but only when an encoder is used. Analyses further reveal that capabilities are transferred through distinct network depths rather than simply stacked, and that a stage converged on its own benchmark is not necessarily ready for the next. Distilled into a simple recipe, these findings yield 4B TS-LLMs that, without any target-domain data, outperform four recent 7B TS-LLMs on seven of eight out-of-domain metrics, by up to 11 points. We therefore advocate treating time series as a first-class modality for LLMs, read through dedicated encoders and taught to perceive before they reason. Code is available at \url{https://github.com/EIT-NLP/PSR}.
\end{abstract}
\section{Introduction}
Time series record how the world evolves, from power grids and patient vitals to markets and traffic, and reading them in context underlies many critical decisions~\citep{jiang2025multi}. In vision and audio, large language models (LLMs) have become general assistants that perceive, describe, and reason about their inputs, built on training and evaluation suites that span these capabilities~\citep{liu2023visual, liu2024mmbench, chu2023qwen, sakshi2025mmau,liu2026vica}. Time series understanding, by contrast, remains underestimated. Most work on time series LLMs (TS-LLMs) applies them to traditional tasks such as forecasting, classification, and anomaly detection~\citep{zhou2023one, ICLR2024_680b2a81, liu2025timecma, kong2025achieving}, reducing understanding to curve fitting~\citep{qiu2026recognition}. Studies of reasoning are few and narrow, often stopping at surface-level question answering~\citep{xie2025chatts, guan2025timeomni}, and benchmarks for generalist models largely overlook time series~\citep{yu2026tsrbench}. Above all, no off-the-shelf suite trains and evaluates the full range of time series capabilities.

We view comprehensive time series understanding as three progressive capabilities: \emph{perceiving} basic attributes such as trend, seasonality, and local patterns; \emph{solving} standard tasks such as forecasting and anomaly detection; and \emph{reasoning} about a series and its context to explain, extrapolate, and decide. Each capability has been studied on its own~\citep{cai2024timeseriesexam, zhou2026cats, hu2025context, liu2026semantic, kong2025time, wang2025itformer, lu2026patra}, and so has the way time series enter the LLM: as text, as plotted images, or through a dedicated encoder with different alignment designs~\citep{liu2025picture, xu2026can, he2026harnessing, xie2025chatts, qiu2026recognition}. These comparisons, however, are each made in a separate setting with its own tasks and data. When one model must acquire all three capabilities, a question arises that isolated settings cannot answer: \textit{how should time series enter the LLM, and in what order should the capabilities be learned?} The answer hinges on whether what is learned for one capability carries over to the next.

To answer this question, we build, to our knowledge, the first suite that trains and evaluates all three capabilities. It curates four public corpora, covering synthetic series and real-world series from over ten domains~\citep{xie2025chatts, kong2025time, quinlan2026chat, qiu2026recognition}. The data are organized into three stages, Perceive, Solve, and Reason, each with its own training set and held-out evaluation set, totaling 337K training and 44K evaluation samples. Three external benchmarks are reserved for out-of-domain testing~\citep{chen2025mtbench, cai2024timeseriesexam, zhou2026cats}. On this suite, we compare textualization, which writes the series as text, with two encoder-based designs, pre-alignment and post-alignment, which align encoder outputs before or within the LLM~\citep{qiu2026recognition, kong2025time}. For each design, we contrast \emph{progressive} training, which learns the three stages in order, with \emph{mixed} training on all the data at once and with reasoning-only training. We further examine which perceptual attributes matter, how gradients evolve under each training scheme, and how representations transfer across layers.

In this setting, \textit{a dedicated encoder proves essential}. Textualization trails both encoder-based designs by 8--10 points in reasoning accuracy, although it fine-tunes the LLM while pre-alignment keeps the LLM frozen. Borrowed formats miss fine-grained structure. Even GPT-4o, reading a series as text or as a plot, scores about half as high as our perception-trained 4B models on local patterns and inter-variable relations. Ablating the perception data shows that these are exactly the attributes reasoning relies on most. Between the two alignment designs, the lighter pre-alignment performs on par with post-alignment. Its encoder can even be frozen after perception with little loss, so \textit{later capabilities form mainly by re-aligning fixed temporal features rather than by refining the encoder}.

Given such an encoder, we reach four findings on how the capabilities form. (1) \textit{The capabilities build on one another.} Perception alone improves all four standard tasks, and perception and task solving strengthen complementary aspects of reasoning: the former chiefly helps read the signal and ground it in context, the latter chiefly helps infer dependencies and choose actions. (2) \textit{Order matters beyond the data, but only with an encoder.} On identical data, learning the stages in order beats mixing them, most of all when reasoning data are scarce: with progressive training, $15\%$ of the reasoning data roughly matches reasoning-only training on $50\%$. Under mixed training, reasoning gradients dominate the updates while aligning poorly with the overall objective, and ordering the stages more than doubles this alignment. Without an encoder, however, the advantage nearly vanishes, an interaction that only a multi-capability setting can reveal. (3) \textit{Capabilities are transferred through distinct network depths rather than simply stacked.} Perception-to-solving transfer dominates the shallower layers and solving-to-reasoning transfer the deeper ones, while direct perception-to-reasoning transfer stays weakest, so reasoning is built through task solving. (4) \textit{Local convergence does not mean readiness.} Each preceding stage saturates on its own benchmark with about $70\%$ of its data, yet training on the rest still benefits the stages that follow, so local convergence alone should not decide when to advance.

These findings translate into a simple recipe: we should encode time series with a dedicated encoder and learn the three stages in order, each on its full data. With only 4B parameters, the resulting TS-LLMs achieve the best result on seven of eight out-of-domain metrics, ahead of 7B TS-LLMs such as TimeOmni-1, ITFormer, ChatTS, and PATRA~\citep{guan2025timeomni, wang2025itformer, xie2025chatts, lu2026patra}.

In summary, our contributions are as follows: (1) \textbf{A comprehensive suite.} We build the first training and evaluation suite that covers perception, task solving, and reasoning for time series. (2) \textbf{How time series should enter the LLM.} In this multi-capability setting, we show that a dedicated encoder is essential, that a pre-aligned encoder trained once suffices, and that later capabilities form by re-alignment. (3) \textbf{How capabilities form.} We show that the capabilities build on one another and are best learned in order, explain this through gradient alignment, depth-wise representation transfer, and stage readiness, and turn the findings into a recipe for strong, compact TS-LLMs.
\section{Related Work}
\textbf{LLM for Time Series Analysis.}\quad
Recent studies have explored the potential of LLMs for time series analysis across various tasks, including forecasting, classification, anomaly detection, and QA~\citep{qiu2025few,kong2025time,wang2025itformer,xu2026can}. These works investigate how the semantic understanding and instruction-following capabilities of LLMs can be adapted to temporal data~\citep{ICLR2024_680b2a81,qiu2026rethinking,he2026harnessing}. For instance, CrisisTS integrates textual crisis reports with meteorological time series to improve multilingual urgency assessment beyond text-only approaches~\citep{meunier2025crisists}. Time-MQA extends LLMs to multi-task temporal QA by combining language reasoning with time series information~\citep{kong2025time}, while ITFormer develops an end-to-end framework that jointly models temporal patterns and language generation for multimodal QA in engine maintenance scenarios~\citep{wang2025itformer}. Existing studies mainly target individual downstream tasks, while general temporal capability formation remains underexplored.

\textbf{QA with Language and Time Series.}\quad
Some studies have explored the integration of language models with time series data to enable interactive temporal QA~\citep{xie2025chatts,kong2025time,wang2025itformer,guan2025timeomni,yu2026tsrbench}. These works construct time series QA datasets and leverage instruction tuning, chain-of-thought reasoning, and alignment strategies to extend TS-LLMs from numerical prediction toward temporal understanding~\citep{xie2025chatts,kong2025time,wang2025itformer,guan2025timeomni}. For example, Time-MQA introduces multi-task temporal QA by combining language reasoning with time series information~\citep{kong2025time}, while recent frameworks further investigate complex temporal analysis through instruction following and reasoning-oriented training~\citep{wang2025itformer,guan2025timeomni,yu2026tsrbench}. Existing approaches mainly optimize QA performance, while the temporal perception and task-solving foundations of reasoning remain underexplored. Direct QA supervision may therefore yield strong responses without grounded temporal understanding.

\section{A Progressive Three-Stage Capability Dimensions for TS-LLMs}
\vspace{-0.3cm}
Despite recent progress in applying LLMs to time series perception, task solving, and reasoning, these capabilities are often studied independently. To analyze the relationships among different tasks, we organize the capabilities of TS-LLMs into three stages: Perceive, Solve, and Reason. More data details are provided in Appendix~\ref{app:Overview} \& \ref{app:Dataset}.

\begin{wrapfigure}{r}{0.6\linewidth}
    \small
    \centering
    \vspace{-12pt}
    \includegraphics[width=\linewidth]{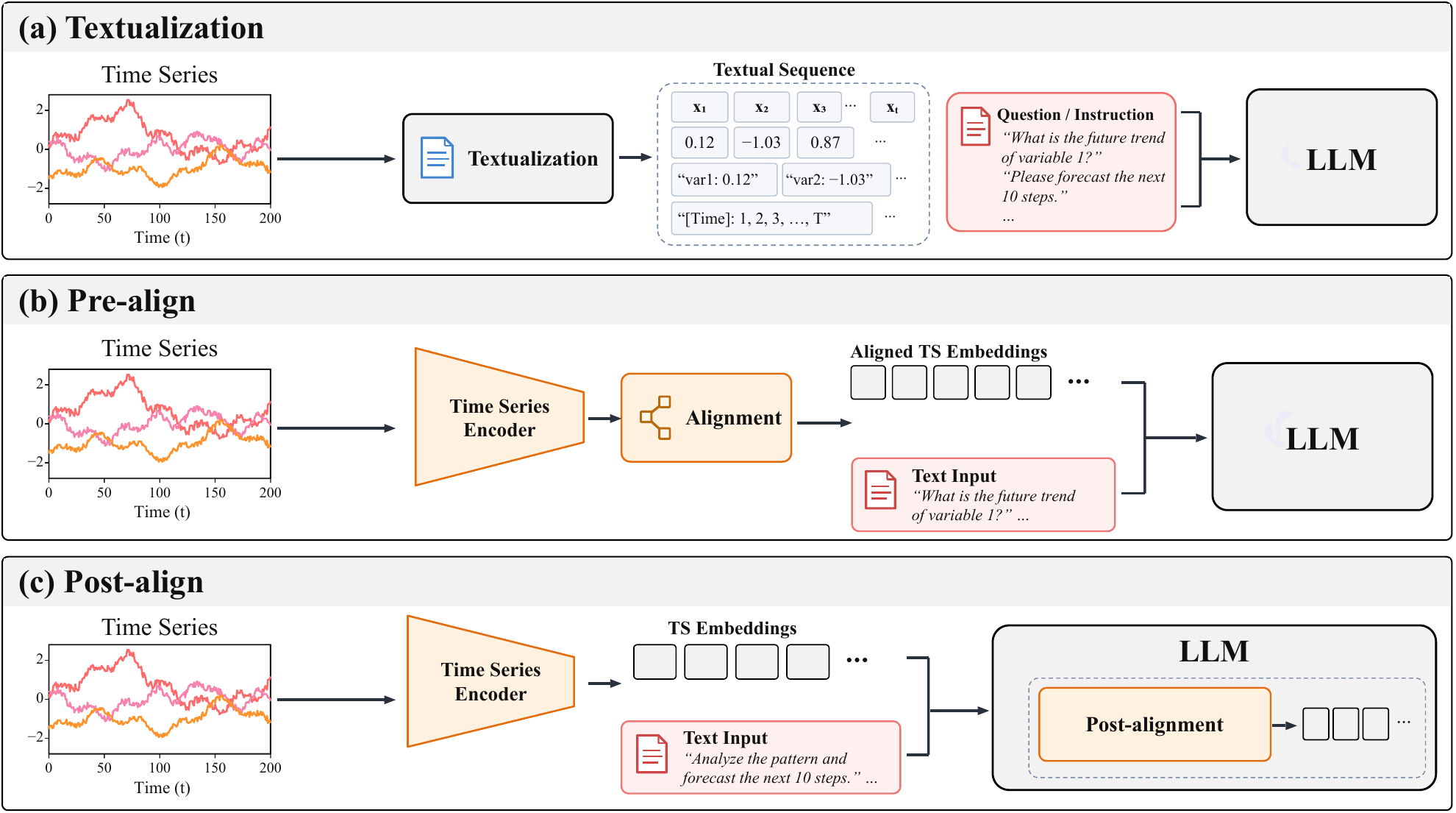}
    % \vspace{-0.6cm}
    \caption{\small Three Model Architecture.}
    \label{fig:architecture}
    \vspace{-12pt}
\end{wrapfigure}

\textbf{Model Architecture.}\quad
Existing studies have developed three main paradigms in Figure~\ref{fig:architecture}. \textit{\textbf{(1) Time Series Textualization}} directly converts time series into textual sequences that can be processed by LLMs. \textit{\textbf{(2) Pre-align}} architecture extracts temporal representations with an independent time series encoder and projects them into the LLM embedding space, enabling joint processing with text inputs. \textit{\textbf{(3) Post-align}} architecture first encodes raw time series into continuous temporal embeddings and feeds them into the LLM as additional inputs. In our experiments, we adopt TSAlign~\citep{qiu2026recognition} and Time-MQA~\citep{kong2025time} as representative Pre-align and Post-align architectures, with Qwen3-4B~\citep{yang2025qwen3} as the LLM backbone. In Pre-align, we train only the encoder and alignment module; in Post-align, we train the encoder and LLM backbone; and in the textualization setting, only the LLM is fine-tuned via SFT.

\textbf{Stage 1: Fundamental Attribute Perception.}\quad
Advanced time series analysis relies on understanding fundamental temporal dynamics. For example, forecasting requires capturing trend patterns, while anomaly analysis requires identifying local behaviors. Therefore, Stage 1 aims to equip LLMs with basic temporal perception. The questions focus on trends, seasonality, anomalies, local patterns, and multivariate relationships.
\textit{\textbf{Training Set.}}\quad
The training data is mainly collected from ChatTS, with approximately $\sim$105K samples~\citep{xie2025chatts}.
\textit{\textbf{Evaluation Set.}}\quad
We use two datasets to assess perception. Dataset A contains 525 constructed questions. Dataset B contains 1,616 questions. The evaluation includes \textit{categorical} and \textit{numerical} tasks. Categorical tasks are evaluated by \textit{F1-score}, while numerical tasks are evaluated by \textit{relative accuracy}.

\textbf{Stage 2: Temporal Task Solving.}\quad
Many real-world time series applications require models to solve specific tasks. Therefore, Stage 2 investigates whether models can transfer basic temporal perception abilities to downstream task solving.
\textit{\textbf{Training Set.}}\quad
We use TSQA~\citep{chen2025mtbench} and TS Instruct~\citep{quinlan2026chat} to train task-solving abilities. TSQA contains 155,214 samples, including 42,557 forecasting, 38,657 imputation, 37,000 anomaly detection, and 37,000 classification samples. TS Instruct provides an additional 2,945 classification samples, resulting in 158,159 samples in total. All samples are split into training and testing sets with an 8:2 ratio.
\textit{\textbf{Evaluation Set.}}\quad
We evaluate four standard tasks. Forecasting and imputation are evaluated using \textit{Mean Squared Error (MSE)}, while anomaly detection and classification are evaluated using \textit{Accuracy}.

\textbf{Stage 3: High-level Reasoning.}\quad
Stage 3 aims to advance TS-LLMs to intelligent systems capable of analysis.
\textit{\textbf{Training Set.}}\quad
TSQA reasoning QA provides 26,163 samples, ChatTS TSEvol and Instruct Follow~\citep{xie2025chatts} provide 29,320 samples, the TSCognition~\citep{qiu2026recognition} training set provides 32,869 samples, and TS Instruct provides 12,852 samples, resulting in 105,751 samples in total.
\textit{\textbf{Evaluation Set.}}\quad
We use two independent sets. Set A contains 4,547 samples covering binary, multiple-choice, and open-ended questions, evaluated using \textit{Accuracy}, \textit{ROUGE-L}, and \textit{BLEU}. Set B contains 6,163 samples and evaluates five dimensions: \textit{Decoding}, \textit{Grounding}, \textit{Inferring}, \textit{Extrapolating}, and \textit{Acting}. The metric is the accuracy of multiple-choice questions.

\section{Investigating Capability Formation in TS-LLMs}
\vspace{-0.3cm}
\label{Investigating Capability Formation in TS-LLMs}
We formulate five research questions (RQs) to investigate the mechanisms underlying capability relationships in TS-LLMs. They examine cross-stage transfer, progressive supervision, input format, stage-wise convergence, and the conditions that support high-level capability formation. The LLM backbone is primarily Qwen3-4B, with additional Qwen3-1.7B results reported in Appendix~\ref{Supplementary}.

\begin{figure}[t]
    \centering
    % \vspace{-0.6cm}
    \includegraphics[width=0.9\linewidth]{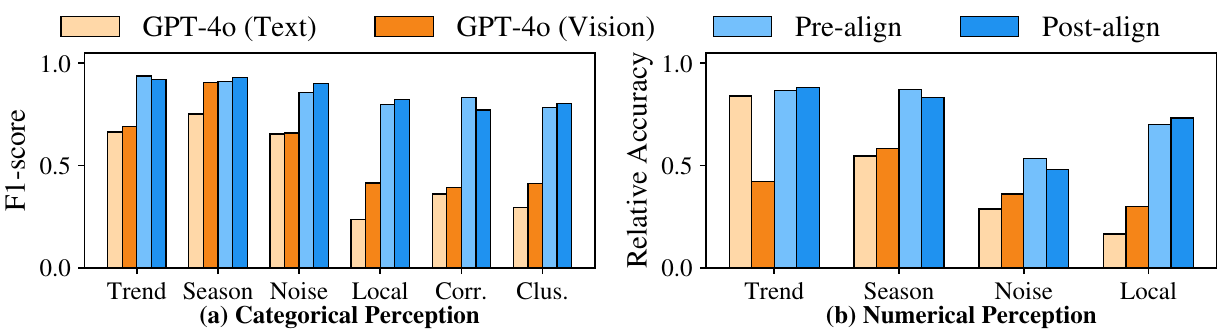}
    % \vspace{-0.3cm}
    \caption{\small Categorical and numerical perception comparison across different methods.}
    \label{fig:cate_num_comparison_four_models}
\end{figure}

\begin{table}[t]
\centering
\resizebox{0.9\linewidth}{!}{
\begin{tabular}{c|l|cccc}
\toprule
\textbf{Model} & \textbf{Training Setting} & 
\textbf{Fore. MSE $\downarrow$} & 
\textbf{Impu. MSE $\downarrow$} & 
\textbf{Ano Det. Acc. $\uparrow$} & 
\textbf{Class. Acc. $\uparrow$} \\
\midrule
\textbf{Qwen3-4B}& No fine-tuning & 1.970 & 0.0286 & 0.333 & 0.296 \\
\midrule
\multirow{3}{*}{\textbf{Pre-align}} 
& Stage 1 only & 1.842 & 0.0194 & 0.587 & 0.459 \\
& Stage 2 only & 1.368 & 0.0136 & 0.682 & 0.567 \\
& Stage (1 $\rightarrow$ 2) & \textbf{1.328} & \textbf{0.0132} & \textbf{0.739} & \textbf{0.628} \\
\midrule
\multirow{3}{*}{\textbf{Post-align}}
& Stage 1 only & 1.716 & 0.0217 & 0.612 & 0.431 \\
& Stage 2 only & 1.423 & \textbf{0.0146} & 0.660 & 0.529 \\
& Stage (1 $\rightarrow$ 2) & \textbf{1.286} & 0.0159 & \textbf{0.720} & \textbf{0.597} \\
\bottomrule
\end{tabular}
}
% \vspace{-0.2cm}
\caption{\small Results of four task solving evaluation sets.}
\label{tab:progressive_capability}
% \vspace{-0.3cm}
\end{table}

\textbf{RQ1: Can Temporal Perception Facilitate to Task Solving?}\quad
We first evaluate the perception ability of models after Stage 1 training. As shown in Figure~\ref{fig:cate_num_comparison_four_models} \& Table~\ref{tab:cate_num_comparison_detailed}, we compare two TS-LLM architectures, Pre-align and Post-align on categorical and numerical tasks, with general multimodal LLMs, GPT-4o (Text) and GPT-4o (Vision), where the GPT-4o results are adopted from ~\citep{xie2025chatts}. Both Pre-align and Post-align TS-LLMs outperform GPT-4o on most tasks, showing that single Stage 1 training enables TS-LLMs to develop effective fundamental perception.
We compare different learning paradigm on the Stage 2 task solving evaluation set to investigate whether perception abilities can facilitate task solving. As shown in Table~\ref{tab:progressive_capability}, \textbf{\textit{Stage 1 only}} denotes training only on the training set of Temporal Perception. \textbf{\textit{Stage 2 only}} directly trains the model on the training set of Task Soving, and \textbf{\textit{Stage (1 $\rightarrow$ 2)}} denotes sequential training first on Temporal Perception and then on Task Solving.
% Results show that Stage 1 training provides effective temporal knowledge for subsequent task solving. Compared with the non-fine-tuned Qwen3-4B, Stage 1 only achieves improvements across all four tasks, indicating that fundamental temporal understanding can transfer to practical time series applications. Moreover, comparing Stage 2 only with Stage (1 $\rightarrow$ 2) reveals that temporal knowledge acquired during the Perceive stage can be transferred to the Solve stage, leading to further improvements in task solving performance.

\textbf{Conclusion: }\textit{(1) Stage 1 provides effective fundamental perception.} Compared with the non-fine-tuned model, Stage 1 improves all tasks, indicating effective acquisition of fundamental temporal attributes.
\textit{(2) Stage 2 is the primary source of performance improvement.} Training on task-solving objectives enhances the corresponding downstream capabilities.
\textit{(3) Temporal perception can effectively facilitate downstream task solving.} Stage (1 $\rightarrow$ 2) obtains the best results on most tasks.

\begin{table}[t]
\centering
% \vspace{-0.5cm}
\resizebox{0.85\linewidth}{!}{
\begin{tabular}{c|l|ccccc}
\toprule
\textbf{Model} & \textbf{Training Setting} &
\textbf{Decoding} &
\textbf{Grounding} &
\textbf{Inferring} &
\textbf{Extrapolating} &
\textbf{Acting} \\
\midrule

\multirow{4}{*}{\textbf{Pre-align}}
& Stage 3 Only
& 72.2 & 68.5 & 61.3 & 57.2 & 50.7 \\

& Stage (1 $\rightarrow$ 3)
& 77.8 & \textbf{81.5} & 63.3 & 62.2 & 51.8 \\

& Stage (2 $\rightarrow$ 3)
& 75.1 & 72.0 & 65.8 & 59.5 & 53.9 \\

& Stage (1 $\rightarrow$ 2 $\rightarrow$ 3)
& \textbf{82.5} & 79.1 & \textbf{70.7} & \textbf{65.4} & \textbf{57.2} \\

\midrule

\multirow{4}{*}{\textbf{Post-align}}
& Stage 3 Only
& 68.5 & 66.2 & 58.9 & 58.8 & 49.2 \\

& Stage (1 $\rightarrow$ 3)
& 75.6 & 73.8 & 61.2 & 64.1 & 54.3 \\

& Stage (2 $\rightarrow$ 3)
& 73.9 & 72.7 & \textbf{66.0} & \textbf{68.9} & 58.8 \\

& Stage (1 $\rightarrow$ 2 $\rightarrow$ 3)
& \textbf{79.0} & \textbf{79.4} & 64.2 & 66.7 & \textbf{61.4} \\

\bottomrule
\end{tabular}
}
% \vspace{-0.2cm}
\caption{\small Accuracy $\uparrow$(\%) of different strategies on the Evaluation Set B of high-level reasoning.}
\label{tab:reasoning_results}
\end{table}

\begin{table}[t]
\centering
\resizebox{0.85\linewidth}{!}{
\begin{tabular}{c|l|cccc}
\toprule
\multirow{2}{*}{\textbf{Model}} &
\multirow{2}{*}{\textbf{Training Setting}} &
\textbf{Ture/False QA} &
\textbf{Multi-Choice QA} &
\textbf{Open-end QA} &
\textbf{Open-end QA} \\
&
&
Acc. $\uparrow$ &
Acc. $\uparrow$ &
ROUGE-L$\uparrow$ &
BLEU$\uparrow$ \\
\midrule

\multirow{4}{*}{\textbf{Pre-align}}
& Stage 3 Only
& 71.4 & 58.6 & 43.8 & 24.7 \\

& Stage (1 $\rightarrow$ 3)
& 78.3 & 62.2 & 45.5 & 25.8 \\

& Stage (2 $\rightarrow$ 3)
& 74.5 & 65.1 & 47.2 & 26.7 \\

& Stage (1 $\rightarrow$ 2 $\rightarrow$ 3)
& \textbf{80.6} & \textbf{68.8} & \textbf{51.6} & \textbf{31.5} \\

\midrule

\multirow{4}{*}{\textbf{Post-align}}
& Stage 3 Only
& 66.3 & 60.3 & 42.1 & 22.3 \\

& Stage (1 $\rightarrow$ 3)
& 70.4 & 63.6 & 40.8 & 24.5 \\

& Stage (2 $\rightarrow$ 3)
& 68.6 & 67.6 & 44.4 & 25.9 \\

& Stage (1 $\rightarrow$ 2 $\rightarrow$ 3)
& \textbf{72.3} & \textbf{71.5} & \textbf{49.7} & \textbf{28.9} \\

\bottomrule
\end{tabular}
}
% \vspace{-0.2cm}
\caption{\small Performance on the Evaluation Set A of high-level reasoning.}
\label{tab:reasoning_path}
\end{table}

\textbf{RQ2: Do Perceive and Solve Facilitate Reason Formation?}\quad
Recent studies have advanced TS-LLMs toward complex QA reasoning~\citep{wang2025itformer,guan2025timeomni}. However, QA supervision mainly optimizes final answers and may overlook critical temporal evidence. We therefore compare four capability development paths across five dimensions.
Experimental results in Table~\ref{tab:reasoning_results} and~\ref{tab:reasoning_path} show that introducing preceding stages improves reasoning performance compared with directly training on Reasoning datasets. On Evaluation Set B, \textbf{\textit{Stage (1 $\rightarrow$ 3)}} achieves higher accuracy than \textbf{\textit{Stage 3 Only}}, indicating that perception helps models better utilize evidence. \textbf{\textit{Stage (2 $\rightarrow$ 3)}} also improves multiple reasoning dimensions, suggesting that task solving experience provides richer temporal dynamics for subsequent reasoning. Similar trends are observed on Evaluation Set A, where both preceding-stage strategies improve True/False, multiple-choice, and open-ended QA performance in most cases. Overall, \textbf{\textit{Stage (1 $\rightarrow$ 2 $\rightarrow$ 3)}} achieves the best results on most metrics.

\textbf{Conclusion: }\textit{(1) Perceive and Solve jointly facilitate the formation of Reason.} Temporal perception provides fundamental knowledge, while task-solving experience exposes models to richer dynamics, together supporting more effective high-level reasoning.
\textit{(2) The complete Stage (1 $\rightarrow$ 2 $\rightarrow$ 3) pathway achieves the best overall performance.} 

\begin{figure}[t]
    \centering
    % \vspace{-0.4cm}
    \includegraphics[width=0.8\linewidth]{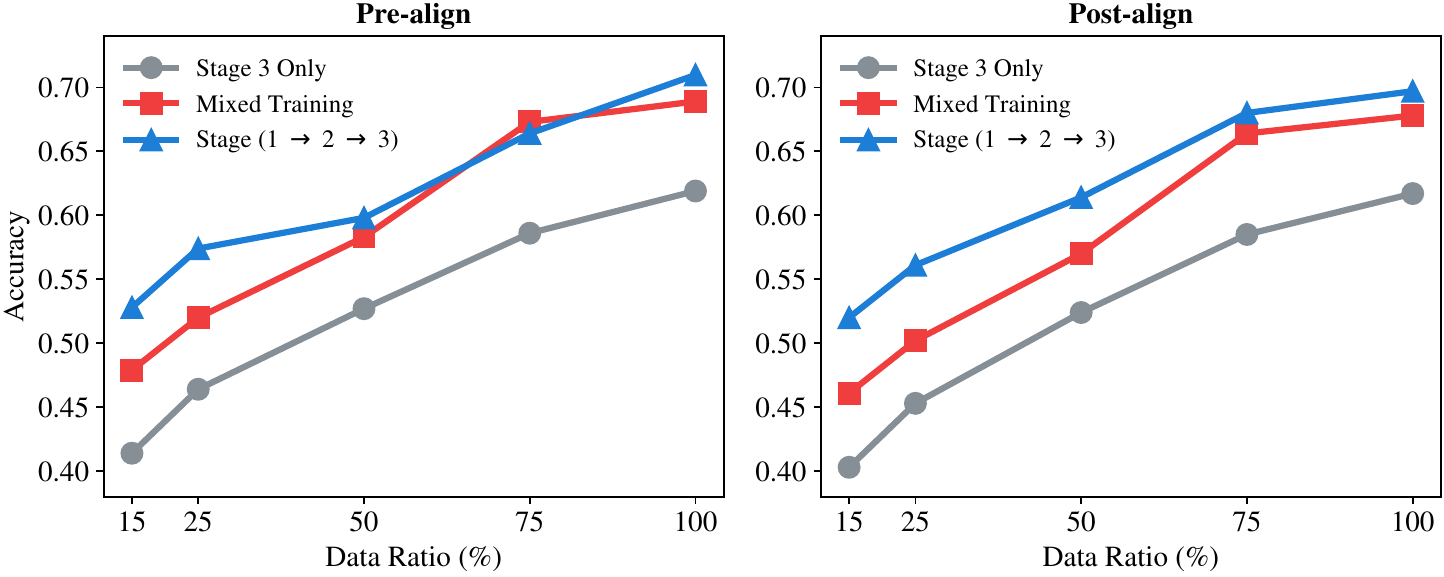}
    % \vspace{-0.2cm}
    \caption{\small Accuracy across different training data scales.}
    \label{fig:reasoning_data_scaling}
\end{figure}

% \begin{wrapfigure}{r}{0.62\linewidth}
%     \small
%     \centering
%     \vspace{-0.5cm}
%     \includegraphics[width=\linewidth]{figure/reasoning_data_scaling.pdf}
%     \vspace{-0.6cm}
%     \caption{\small Accuracy across different training data scales.}
%     \label{fig:reasoning_data_scaling}
%     \vspace{-0.5cm}
% \end{wrapfigure}

\textbf{RQ3: Does Reason Benefit from Progressive Capability Accumulation?}\quad
RQ2 has shown that both Perceive and Solve provide effective foundations for Reason. We further investigate whether these capabilities need to be accumulated progressively following the stage order. To this end, we compare three training strategies: \textbf{\textit{Stage 3 Only}}, \textbf{\textit{Mixed Training}}, and \textbf{\textit{Stage (1$\rightarrow$ 2 $\rightarrow$ 3)}}. We gradually vary the proportion of high-level reasoning data from 15\% to 100\%, while keeping the perception and task-solving training data unchanged.
Figure~\ref{fig:reasoning_data_scaling} \& Table~\ref{tab:reasoning_data_scaling_detailed} shows that progressive capability accumulation achieves stronger Reason performance on the Evaluation Set B of high-level reasoning across most data scales. When only 15\% of Stage 3 data is used, \textbf{\textit{Stage (1$\rightarrow$2$\rightarrow$3)}} reaches 0.528 and 0.520 on Pre-align and Post-align, clearly outperforming both \textbf{\textit{Mixed Training}} and \textbf{\textit{Stage 3 Only}}. As more Stage 3 data is introduced, all strategies improve, while the progressive pathway remains highly competitive.

\textbf{Conclusion: }\textit{(1) Ordered capability accumulation is more effective than mixed supervision for Reason formation.} Under the same supervision sources, progressive training consistently outperforms mixed training. 
\textit{(2) Progressive capability accumulation improves Reason supervision efficiency.} Its advantage becomes more pronounced when Stage 3 data is limited.

\begin{wrapfigure}{r}{0.50\linewidth}
    \small
    \centering
    \vspace{-0.2cm}
    \includegraphics[width=\linewidth]{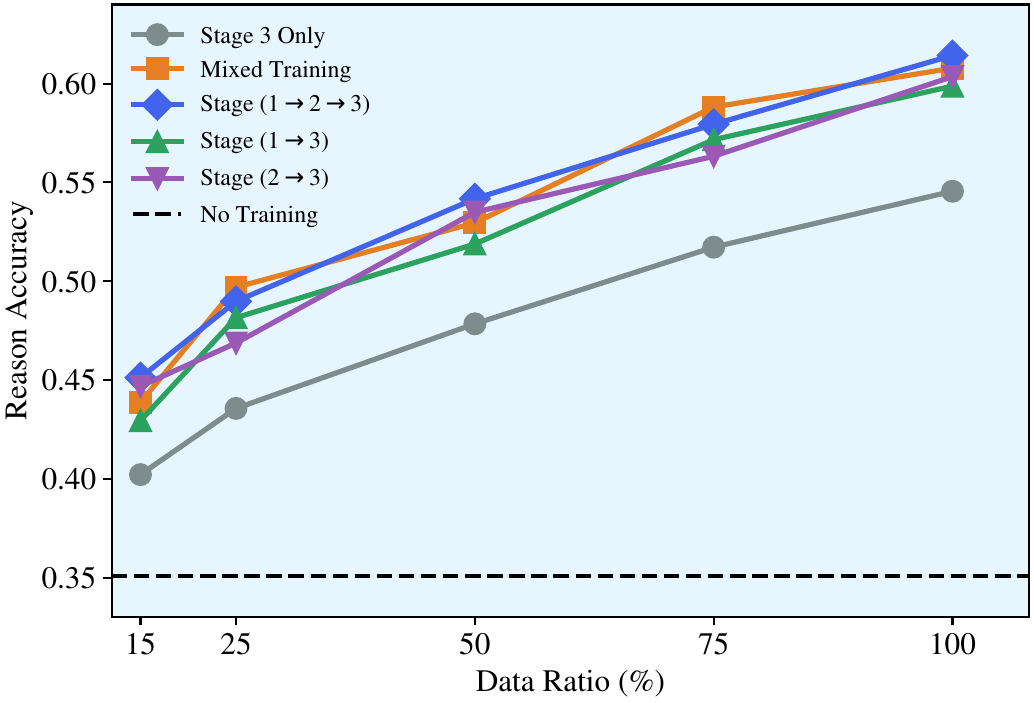}
    \vspace{-0.6cm}
    \caption{\small Reason under the Textualization paradigm across training data scales of high-level reasoning.}
    \label{fig:direct}
    \vspace{-0.3cm}
\end{wrapfigure}
\textbf{RQ4: Does Time Series Input Format Affect Progressive Formation?}\quad
We adopt the \textit{Textualization} paradigm, which directly feeds time series representations into the LLM, and compare \textbf{\textit{Stage 3 Only}}, \textbf{\textit{Mixed Training}}, \textbf{\textit{Stage (1 $\rightarrow$ 3)}}, \textbf{\textit{Stage (2 $\rightarrow$ 3)}}, and \textbf{\textit{Stage (1 $\rightarrow$ 2 $\rightarrow$ 3)}} under the same training paths. Reason performance is evaluated on the Evaluation Set A of high-levle reasoning. Results are shown in Figure~\ref{fig:direct} \& Table~\ref{tab:reasoning_data_scaling_detailed}. \textbf{\textit{No Training}} denotes direct evaluation of the original Qwen3-4B without any additional training.
We observe that under the Textualization paradigm, all training strategies improve steadily as more training data is introduced. However, the gaps between different capability paths are much smaller than those under Pre-align and Post-align. In addition, the complete \textbf{\textit{Stage (1 $\rightarrow$ 2 $\rightarrow$ 3)}} pathway provides only limited gains and performs similarly to \textbf{\textit{Mixed Training}}, \textbf{\textit{Stage (1 $\rightarrow$ 3)}}, and \textbf{\textit{Stage (2 $\rightarrow$ 3)}}. Our results indicate that progressive capability formation is more evident in TS-LLMs with dedicated cross-modal alignment, while the cross-stage accumulation effect is weakened under direct textualized input.

\textbf{Conclusion: }\textit{(1) Time series input format affects progressive capability formation.} 
Input strategies with temporal modeling exhibit clearer progressive gains, while direct textualization weakens this effect.
\textit{(2) Explicit temporal integration facilitates cross-stage capability transfer.} 
It better preserves and transfers preceding capabilities across progressive stages.

\begin{table}[t]
\centering
% \vspace{-0.4cm}
\resizebox{0.99\linewidth}{!}{
\begin{tabular}{l|c|c|c|c}
\toprule
\textbf{Training Setting} &
\textbf{Training Data} &
\textbf{Evaluation Performance} &
\textbf{Relative Perf.} &
\textbf{I$\rightarrow$II$\rightarrow$III} \\
\midrule

Stage 1 only
& (40\% $\rightarrow$ 100\%) I 
& 0.613, 0.662, \textbf{0.738}, 0.745, 0.751
& 98.3\%
& 0.743 \\

\midrule

Stage (1 $\rightarrow$ 2)
& 70\% I, (40\% $\rightarrow$ 100\%) II
& 0.542, 0.603, \textbf{0.644}, 0.649, 0.652
& 98.8\%
& 0.684 \\

\midrule

Stage (1 $\rightarrow$ 2 $\rightarrow$ 3)
& 70\% I, 70\% II, 
 (40\% $\rightarrow$ 100\%) III
& 0.537, 0.569, 0.616, \textbf{0.663}, 0.682
& 97.2\%
& 0.710 \\

\bottomrule
\end{tabular}
}
% \vspace{-0.2cm}
\caption{\small Stage-wise performance of the Pre-align model under increasing training data scales.
For each stage, we progressively increase the training data while fixing preceding stages at their identified saturation points. Stage 1 uses Numerical Perception Relative Accuracy; Stage 2 uses Anomaly Detection and Classification Accuracy; Stage 3 uses the average Accuracy on Evaluation Set B. For convenient evaluation, we select accuracy-based tasks as  measures. \textbf{Bold} denotes the local saturation points.}
\label{tab:stage_saturation}
\end{table}

\textbf{RQ5: Does Stage-wise Convergence Ensure TS-LLMs Progression?}\quad
During progressive learning, a key question is whether convergence at a preceding stage provides a sufficiently stable foundation for subsequent capability development. Similar to how pretraining states affect post-training in LLMs, local convergence does not necessarily imply sufficient preparation for later capability formation~\citep{akter2026front,chen2026coverage}. We therefore examine two processes: \textbf{\textit{(i) when increasing data in a preceding stage causes its stage-specific performance to enter a saturation regime;}} (ii) \textbf{\textit{whether additional data from that stage can still provide gains for subsequent capability formation after its local performance has largely stabilized.}}
As shown in Table~\ref{tab:stage_saturation} \& Table~\ref{tab:stage_saturation_detailed}, local performance at each stage reaches a clear saturation regime before using the full training set. \textbf{\textit{Stage 1 only}} and \textbf{\textit{Stage (1$\rightarrow$2)}} achieve 98.3\% and 98.8\% of their full-stage performance at around 70\% data, while \textbf{\textit{Stage (1$\rightarrow$2$\rightarrow$3)}} reaches 97.2\% at 85\%. This indicates that each stage appears nearly converged under its own evaluation. However, local saturation does not imply that the model is fully prepared for subsequent capability development. Full I$\rightarrow$II$\rightarrow$III training still yields additional gains, especially for Solve and Reason.

\textbf{Conclusion: }\textit{Stage-wise convergence does not necessarily indicate capability sufficiency for subsequent progression.} A capability can approach its local performance ceiling while additional training at that stage still provides valuable foundations for later capability formation, suggesting that local saturation alone is insufficient to determine when progressive training should advance.

\section{Mechanisms of Progressive Capability Formation}
To understand why progressive learning facilitates capability development, we examine the internal mechanisms underlying the three-stage pathway. We conduct analyses from the perspectives of optimization dynamics (Sec.~\ref{sec:Optimization Dynamics}), temporal attribute contribution (Sec.~\ref{sec:Impact of Temporal Attributes on Reasoning Capability}), representation transfer across stages(Sec.~\ref{sec:Representation Transfer across Stages}), and alignment-driven capability formation (Sec.~\ref{sec:Alignment-Driven Capability Formation}).

\subsection{Optimization Dynamics}
\label{sec:Optimization Dynamics}
To understand the optimization mechanism behind progressive capability formation, we analyze the training dynamics from the perspective of data-induced gradient evolution.

\textbf{Formulation.}
\textit{We consider 3 capability stages, with data distributions denoted as $\mathcal{D}_{P}$, $\mathcal{D}_{S}$, and $\mathcal{D}_{R}$. All stages share the same objective:
$
\ell(\theta;x,y)=-\sum_{t}\log p_{\theta}(y_t|x,y_{<t}).
$
For each capability stage $i\in\{P,S,R\}$, the corresponding optimization objective is defined over its data distribution:
$
\mathcal{L}_{i}(\theta)=
\mathbb{E}_{(x,y)\sim\mathcal{D}_{i}}
[\ell(\theta;x,y)].
$
The gradient induced by each capability objective is given by:
$
g_i(\theta)=\nabla_{\theta}\mathcal{L}_{i}(\theta).
$
Under Mixed Training, multiple capability objectives are optimized simultaneously~\citep{sener2018multi}. The overall update direction is determined by the aggregated gradient:
$
g_{\mathrm{mix}}(\theta)=
\sum_{i\in\{P,S,R\}}\alpha_i g_i(\theta),
$
where $\alpha_i$ denotes the sampling proportion of each stage. Balancing gradients from different objectives is important for stable multi-task optimization~\citep{chen2018gradnorm}. Such joint optimization may introduce competition among capability objectives, causing different stages to follow inconsistent optimization trajectories~\citep{yu2020gradient,liu2021conflict}.
Progressive Training follows the ordered capability pathway:
$
P\rightarrow S\rightarrow R,
$
where each stage is optimized sequentially. This separates the optimization process and provides stage-specific updates during capability formation.}

\textbf{Definition.}
\textit{\textbf{(1) Optimization Strength.} For each capability stage $i$, the optimization strength is measured by the squared gradient norm:
$
G_i(\theta)=\|g_i(\theta)\|_2^2,
$
where $g_i(\theta)$ denotes the gradient induced by stage $i$ data. A larger $G_i(\theta)$ indicates stronger optimization signals from the corresponding capability objective. Under mixed training, the relative gradient contribution is computed as
$
R_i={G_i(\theta)}/{\sum_j G_j(\theta)}.
$
\textbf{(2) Optimization Alignment.}
The directional compatibility between capability-specific optimization and the overall training objective is quantified by:
$
A_i={g_i^Tg_{\mathrm{all}}}/{\|g_i\|_2\|g_{\mathrm{all}}\|_2},
$
where $g_{\mathrm{all}}=\sum_{i\in\{P,S,R\}}g_i$ denotes the aggregated gradient across all capability stages. A larger $A_i$ indicates that the capability-specific gradient follows the global optimization direction more closely.
\textbf{(3) Gradient Update Efficiency.}
The efficiency of converting gradients from each capability stage into parameter updates is:
$
U_i=\frac{\|\theta_{t+\Delta}-\theta_t\|_2}{\sum_{k=t}^{t+\Delta}\|g_i(k)\|_2},
$
where $\theta_{t+\Delta}-\theta_t$ denotes the accumulated parameter update during optimization. A larger $U_i$ indicates that gradients from a capability stage are converted into parameter updates more effectively. We further compare the relative update efficiency across capability stages by normalizing $U_i$ with respect to the overall optimization efficiency:
$
\Delta U_i=\frac{U_i}{\sum_j U_j}\times100\%.
$}

\begin{figure}[t]
    \centering
    % \vspace{-0.4cm}
    \includegraphics[width=1\linewidth]{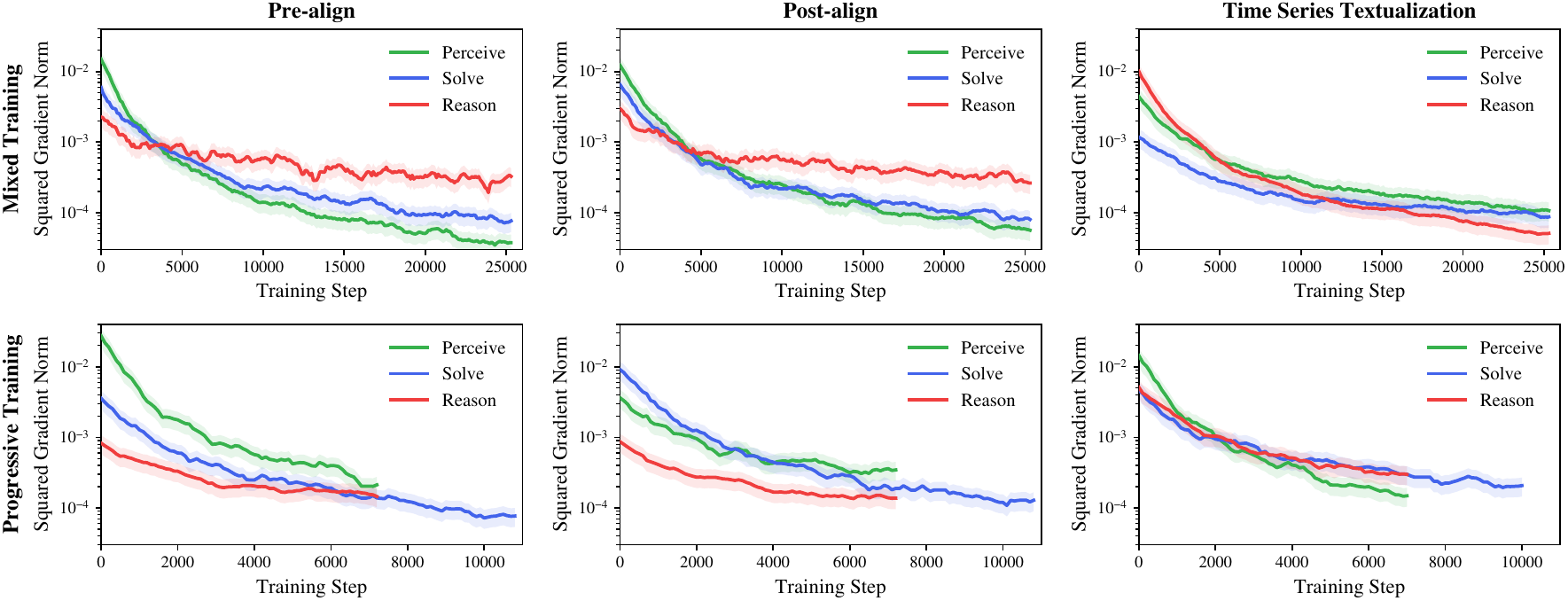}
    % \vspace{-0.2cm}
    \caption{\small Optimization strength dynamics across capability stages.}
    \label{fig:gradient_dynamics_2x3}
\end{figure}

\begin{figure}[t]
    \centering

    % Left: Table
    \begin{minipage}{0.53\linewidth}
        \centering
        \renewcommand{\arraystretch}{1.0}
        \resizebox{\linewidth}{!}{
        \begin{tabular}{ccccc}
        \toprule
        \textbf{Model} & \textbf{Setting} & \textbf{Perceive} & \textbf{Solve} & \textbf{Reason} \\
        \midrule
        \multirow{2}{*}{Pre-align} 
        & Mixed       & 0.463 & 0.382 & 0.214 \\
        & Progressive & 0.612 & 0.574 & 0.593 \\
        \midrule
        \multirow{2}{*}{Post-align} 
        & Mixed       & 0.421 & 0.347 & 0.236 \\
        & Progressive & 0.534 & 0.536 & 0.565 \\
        \midrule
        \multirow{2}{*}{TS Textualization} 
        & Mixed       & 0.318 & 0.219 & 0.241 \\
        & Progressive & 0.301 & 0.254 & 0.226 \\
        \bottomrule
        \end{tabular}
        }
        \captionof{table}{\small Optimization alignment between specific and overall optimization directions.}
        \label{tab:optimization_alignment}
    \end{minipage}
    \hspace{0.03\linewidth}
    % Right: Figure
    \begin{minipage}{0.42\linewidth}
        \centering
        \includegraphics[width=\linewidth]{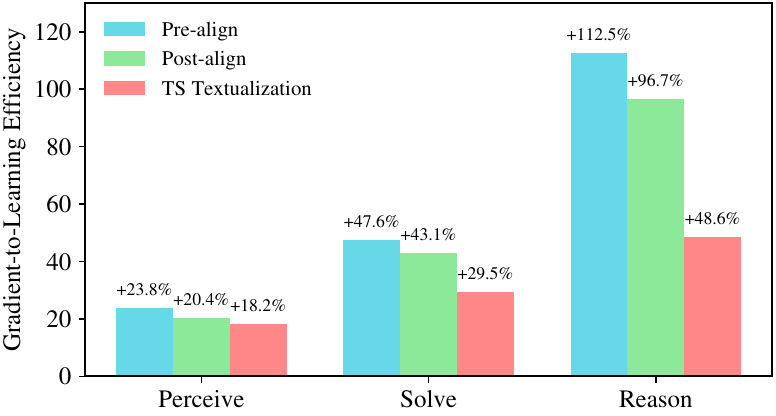}
        \vspace{-0.6cm}
        \captionof{figure}{\small Gradient update efficiency.}
        \label{fig:gradient_learning_efficiency_gain}
    \end{minipage}

\end{figure}

\textbf{Analysis.} As shown in Figure~\ref{fig:gradient_dynamics_2x3}, Optimization Strength reveals distinct optimization dynamics between Mixed Training and progressive training. We quantify the optimization strength of each capability stage using the squared gradient norm. Mixed Training jointly optimizes multiple capability objectives, leading to imbalanced optimization signals, where Reason supervision may dominate parameter updates. In contrast, progressive training allocates optimization strength according to the capability development order, allowing each stage to receive more concentrated optimization signals. TS Textualization fails to achieve optimization dynamics comparable to architectures with Pre-align and Post-align. Therefore, structured temporal representations provide more stable optimization signals.
Building on the analysis of optimization signal magnitude, we compute the alignment score of each capability stage during training. This metric measures the directional consistency between capability-specific gradients and the aggregated gradient, providing a comparison of optimization directions. As shown in Table~\ref{tab:optimization_alignment}, Progressive Training improves gradient alignment under both Pre-align and Post-align architectures. The improvement is particularly evident in the Reason stage, suggesting that progressive training enables high-level capability optimization.
Finally, we compute the relative efficiency gain of Progressive Training over Mixed Training to evaluate how effectively optimization signals are utilized under different temporal representations. As shown in Figure~\ref{fig:gradient_learning_efficiency_gain}, Progressive Training improves optimization efficiency across all temporal representation paradigms. The gain is most significant in the Reason stage, where Pre-align and Post-align achieve efficiency improvements of 112.5\% and 96.7\%, and TS Textualization achieves only 48.6\%.

\subsection{Impact of Temporal Attributes on Reasoning Capability}
\label{sec:Impact of Temporal Attributes on Reasoning Capability}
To analyze the role of perception abilities, we conduct an ablation study over six Stage 1 attributes: \textit{Trend}, \textit{Seasonality}, \textit{Noise}, \textit{Local Pattern}, \textit{Correlation}, and \textit{Clustering}. Specifically, we remove one attribute category at a time while keeping the same three-stage progressive strategy. The final models are evaluated on Evaluation Set A of high-level reasoning to measure their Reason performance.

\begin{figure}[t]
    \centering
    \begin{minipage}{0.45\linewidth}
        \centering
        \includegraphics[width=\linewidth]{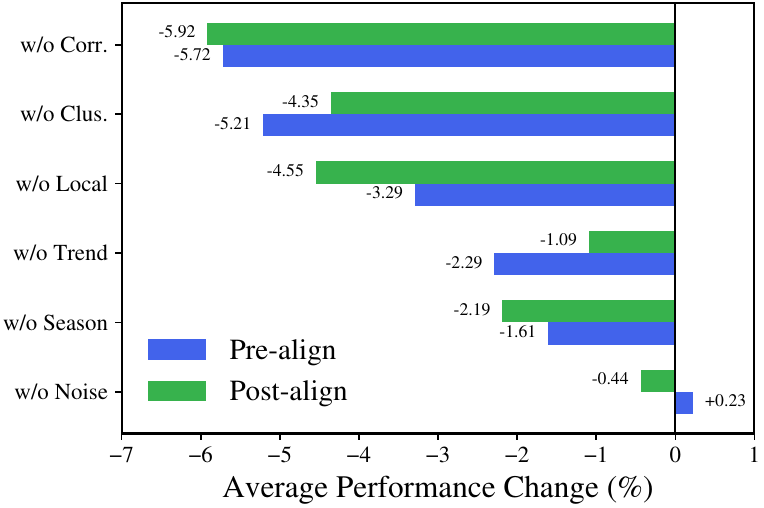}
        \caption{\small Temporal attribute ablation analysis.}
        \label{fig:temporal_attribute_ablation_average}
    \end{minipage}
    \hspace{0.01\linewidth}
    \begin{minipage}{0.45\linewidth}
        \centering
        \includegraphics[width=\linewidth]{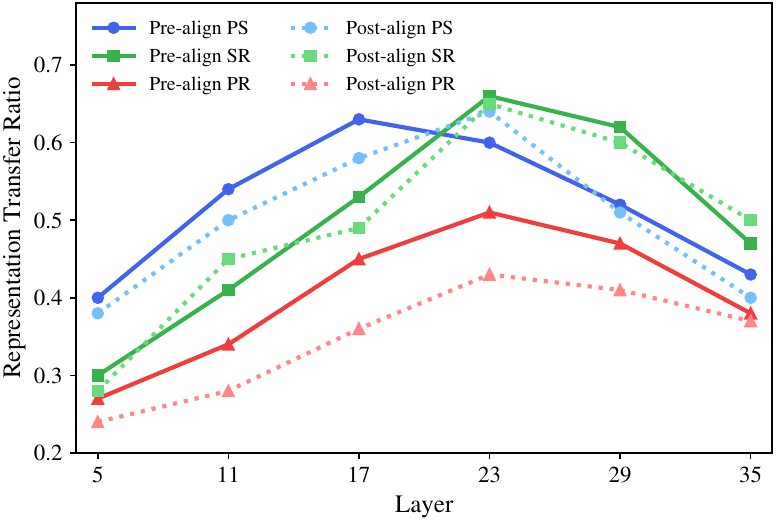}
        \caption{\small Representation transfer across layers.}
        \label{fig:capability_reuse_ratio_layers}
    \end{minipage}
\end{figure}

As shown in Figure~\ref{fig:temporal_attribute_ablation_average}, removing different attributes from Stage 1 leads to varying impacts on the final Reason capability. Among them, removing Correlation, Local Pattern, and Clustering causes the most significant performance degradation. In contrast, Trend and Seasonality have relatively smaller impacts, but still contribute to specific reasoning dimensions. In contrast, removing Noise causes little  degradation, showing that Reason capability relies less on noise-related temporal patterns.

\subsection{Alignment-Driven Capability Formation}
\label{sec:Alignment-Driven Capability Formation}
\begin{wrapfigure}{r}{0.35\linewidth}
    \centering
    \vspace{-0.0cm}
\renewcommand{\arraystretch}{0.8}
    \resizebox{1\linewidth}{!}{
    \begin{tabular}{lcc}
    \toprule
    \textbf{Metric} & \textbf{Non Frozen} & \textbf{Frozen} \\
    \midrule
    Fore. MSE $\downarrow$ & 1.328 & 1.336 \\
    Impu. MSE $\downarrow$ & 0.0132 & 0.0142 \\
    Ano. Acc. $\uparrow$ & 0.739 & 0.718 \\
    Class. Acc. $\uparrow$ & 0.628 & 0.643 \\
    \bottomrule
    \end{tabular}
    }

    \vspace{-0.15cm}
    \captionof{table}{\small Effect of freezing the encoder in task solving evaluation.}
    \label{tab:frozen_encoder}

    \includegraphics[width=1\linewidth]{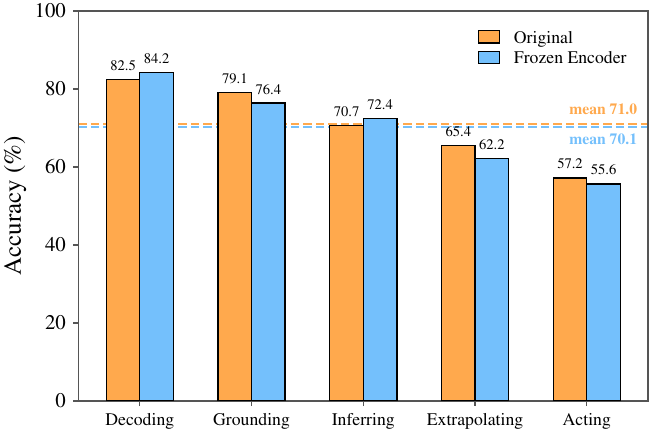}

    \vspace{-0.25cm}
    \captionof{figure}{\small  Effect of freezing the encoder in reasoning evaluation.}
    \label{fig:pre_align_frozen_encoder_stage3}

    \vspace{-0.6cm}
\end{wrapfigure}

Under the Pre-align architecture, the Temporal Encoder is updated only during Stage 1 and then frozen throughout Stages 2 and 3, while only the input-side alignment modules continue to be optimized. Compared with full progressive training, freezing the encoder causes only limited degradation on task solving in Table~\ref{tab:frozen_encoder}, with forecasting, imputation, and anomaly detection showing only small changes and classification remaining comparable. A similar pattern is observed for high-level reasoning in Figure~\ref{fig:pre_align_frozen_encoder_stage3}, where the performance gap between frozen and non-frozen settings remains small across capability stages. 
Our observation suggests that Stage 1 already provides a sufficiently mature temporal feature space to support subsequent capability development. The P $\rightarrow$ S$\rightarrow$ R formation therefore appears to rely mainly on how fixed temporal features are progressively aligned and reorganized for complex downstream capabilities. In this sense, later-stage capability growth is driven more by the evolution of the alignment process than by further modification of the underlying temporal encoder.

\subsection{Representation Transfer across Stages}
\label{sec:Representation Transfer across Stages}
We further investigate capability transfer from the representation perspective. Following the three-stage learning process $P\rightarrow S\rightarrow R$, we adopt the representation projection analysis in~\citep{ke2021achieving,zhao2023rethinking} to measure the proportion of later-stage representations preserved in the representation space of preceding stages. This transfer ratio quantifies the information transmission between stages, where a higher value indicates stronger representation retention across stages.

We conduct a cross-layer analysis on Qwen3-4B to examine representation transfer patterns. We explore three capability transitions: Perceive-to-Solve (PS), Solve-to-Reason (SR), and Perceive-to-Reason (PR). As shown in Figure~\ref{fig:capability_reuse_ratio_layers}, different capability transitions exhibit distinct hierarchical transfer patterns. PS transfer mainly occurs in intermediate layers, suggesting that fundamental temporal structures are transformed into task-solving representations within middle layers. In contrast, SR transfer reaches its peak in deeper layers, indicating that Reason primarily inherits high-level task-oriented representations. PR transfer remains relatively limited, suggesting that advanced reasoning is more likely constructed through the intermediate Solve stage rather than direct capability transfer. Our findings reveal the representation evolution mechanism of Progressive Training: \textbf{\textit{different capabilities are not simply stacked together but are progressively transformed across different network depths, forming a three-stage representation development pathway}}.

\section{Application: Train a Powerful TS-LLM}
% \vspace{-0.3cm}
Based on the above analysis, we adopt a three-stage paradigm, Perceive $\rightarrow$ Solve $\rightarrow$ Reason, to train \textbf{TS-LLM (Pre-align)} and \textbf{TS-LLM (Post-align)}. Similarly, Qwen3-4B is used as the backbone. Stage 1 uses about 105K training samples to learn fundamental temporal attributes, including trends, seasonality, anomalies, local patterns, and multivariate relationships. Stage 2 is built on 158,159 samples, split 8:2 for training and testing, covering forecasting, imputation, anomaly detection, and classification. Stage 3 introduces multi-source data to develop evidence-based temporal analysis. For the \textbf{Pre-align} architecture, the Encoder is updated only in Stage 1 and frozen in Stages 2 and 3. Finally, we obtain two powerful TS-LLMs. 

We conduct out-of-domain (OOD) experiments on MTBench~\citep{chen2025mtbench} while ensuring that no evaluation data are exposed during training. We compare our models with representative TS-LLMs, including ITFormer, ChatTS, TimeOmni-1, and PATRA~\citep{wang2025itformer,xie2025chatts,guan2025timeomni,lu2026patra}. More detailed descriptions of the baselines and datasets are provided in Appendix~\ref{app:Baseline} \& \ref{app:Dataset}. As shown in Table~\ref{tab:ood_results}, our two TS-LLMs achieve the best or second-best results on most metrics across both Weather and Finance domains. TS-LLM (Pre-align) performs particularly well on Weather MCQA, weather prediction, and 7-day finance trend prediction, while TS-LLM (Post-align) achieves the strongest results on 7-day finance MSE and most 30-day finance metrics. In addition, we evaluate temporal perception on TimeSeriesExam~\citep{cai2024timeseriesexam} and CaTS-Bench~\citep{zhou2026cats}. As shown in Table~\ref{tab:perception_generalization}, TS-LLM (Pre-align) achieves the best performance on most metrics.
our results show that the proposed progressive training strategy generalizes effectively to unseen domains and maintains strong performance across different tasks.

\begin{table*}[t]
\centering
% \vspace{-0.4cm}
\small
\resizebox{0.92\textwidth}{!}{
\begin{tabular}{l|cc|ccc|ccc}
\toprule
\multirow{2}{*}{\textbf{Models}} &
\multicolumn{2}{c|}{\textbf{Weather}} &
\multicolumn{3}{c|}{\textbf{Finance (7 days)}} &
\multicolumn{3}{c}{\textbf{Finance (30 days)}} \\
\cmidrule(lr){2-3}
\cmidrule(lr){4-6}
\cmidrule(lr){7-9}
&
\textbf{MCQ. $\uparrow$} &
\textbf{Pre. $\uparrow$} &
\textbf{3-way $\uparrow$} &
\textbf{5-way $\uparrow$} &
\textbf{MSE $\downarrow$} &
\textbf{3-way $\uparrow$} &
\textbf{5-way $\uparrow$} &
\textbf{MSE $\downarrow$} \\
\midrule

TimeOmni-1-7B
& 41.67 & 51.40 & 40.76 & 34.97 & 0.622 & 37.39 & 20.00 & 1.490 \\

ITFormer-7B
& 57.14 & 36.26 & 40.14 & 33.57 & \textcolor{blue}{0.174} & 51.03 & 31.72 & 0.907 \\

ChatTS-7B
& 56.21 & 28.79 & 19.71 & 16.09 & 2.047 & 19.38 & 13.46 & 3.342 \\

PATRA-7B
& 60.46 & 52.04 & \textcolor{blue}{63.33} & \textcolor{red}{60.80} & 0.191 & 54.39 & \textcolor{blue}{43.70} & \textcolor{blue}{0.822} \\

\midrule

\textbf{TS-LLM (Pre-align)}
& \textcolor{red}{67.24}
& \textcolor{red}{57.17}
& \textcolor{red}{68.19}
& \textcolor{blue}{57.82}
& 0.187
& \textcolor{blue}{60.29}
& 41.51
& 0.837 \\

\textbf{TS-LLM (Post-align)}
& \textcolor{blue}{65.92}
& \textcolor{blue}{52.76}
& 59.88
& {55.43}
& \textcolor{red}{0.160}
& \textcolor{red}{65.78}
& \textcolor{red}{47.16}
& \textcolor{red}{0.744} \\

\bottomrule
\end{tabular}
}
% \vspace{-0.3cm}
\caption{\small Out-of-domain results on reasoning tasks. \textcolor{red}{Red}/\textcolor{blue}{blue} indicate the 1st/2nd best results.}
\label{tab:ood_results}
\end{table*}

\begin{table*}[t]
\centering
\small
\resizebox{0.99\textwidth}{!}{
\begin{tabular}{l|ccccc|cccc}
\toprule
\multirow{2}{*}{\textbf{Method}} &
\multicolumn{5}{c|}{\textbf{TimeSeriesExam}} &
\multicolumn{4}{c}{\textbf{CaTS-Bench}} \\
\cmidrule(lr){2-6}
\cmidrule(lr){7-10}
&
\textbf{Ano. Det.} &
\textbf{Causality} &
\textbf{Noise} &
\textbf{Pattern} &
\textbf{Similarity} &
\textbf{Caption Ret.} &
\textbf{TS Ret.} &
\textbf{Amp. Comp.} &
\textbf{Mean Comp.} \\
\midrule

Original Paper
& 28.0 & \textbf{52.0} & 26.0 & 47.0 & 45.0
& 44.0 & 29.0 & 43.0 & 35.0 \\

TimeOmni-1
& 42.6 & 37.5 & 56.0 & 56.4 & 46.7
& \textbf{57.0} & 37.0 & \textbf{92.5} & 67.5 \\

\midrule

\textbf{TS-LLM (Pre-align)}
& \textbf{49.1} & 47.2 & \textbf{60.7} & 61.6 & \textbf{55.0}
& 52.0 & \textbf{46.0} & 82.5 & \textbf{77.5} \\

\bottomrule
\end{tabular}
}
% \vspace{-0.3cm}
\caption{\small Results on TimeSeriesExam and CaTS-Bench. \textbf{Bold} indicates the best results.}
% \vspace{-0.4cm}
\label{tab:perception_generalization}
\end{table*}
\section{Conclusion}
Our work investigates how temporal capabilities progressively form in TS-LLMs. We organize model abilities into three stages: Perceive, Solve, and Reason, and show that ordered capability accumulation provides a stronger foundation for high-level reasoning than isolated or mixed training. Further analyses reveal that capability formation is influenced by training order, input format, stage maturity, and cross-layer representation transfer. We also find that local stage convergence does not necessarily indicate sufficient preparation for subsequent capability development. These observations provide a clearer understanding of how lower-level temporal abilities support more advanced reasoning. Based on these findings, we develop a progressive training strategy for TS-LLMs, which achieves strong out-of-domain performance and offers a practical pathway for building more comprehensive temporal reasoning models.

% \section*{Acknowledgments}
% Add funding information and acknowledgments here, or remove this section.

\bibliographystyle{plainnat}
\bibliography{references}

\clearpage
\appendix

\section{Statistics Overview}
\label{app:Overview}
\textbf{Stage 1: Fundamental Perception.}\quad
Table~\ref{tab:stage1_dataset} summarizes the data used for temporal perception. The training set contains approximately 105K samples from ChatTS, while two independent evaluation sets contain 525 manually constructed questions and 1,616 synthetic questions, respectively. The evaluation covers both categorical and numerical perception tasks.

\begin{table}[h]
\centering
\small
\begin{tabular}{l|l|c|l|l}
\toprule
\textbf{Split} &
\textbf{Source} &
\textbf{Samples} &
\textbf{Task Type} &
\textbf{Metric} \\
\midrule

Training
& ChatTS
& $\sim$105K
& Temporal Perception
& -- \\

\midrule

Evaluation A
& Manual
& 525
& Categorical / Numerical
& F1 / Relative Acc. \\

Evaluation B
& Synthetic
& 1,616
& Categorical / Numerical
& F1 / Relative Acc. \\

\bottomrule
\end{tabular}
\caption{\small Statistics of the Stage 1 perception datasets.}
\label{tab:stage1_dataset}
\end{table}

\textbf{Stage 2: Temporal Task Solving.}\quad
Table~\ref{tab:stage2_dataset} reports the detailed statistics of the task-solving data. The dataset covers forecasting, imputation, anomaly detection, and classification, with 158,159 samples in total. All samples are split into training and testing sets with an 8:2 ratio, and task-specific metrics are used for evaluation.

\begin{table*}[h]
\centering
\small
\resizebox{0.98\textwidth}{!}{
\begin{tabular}{l|l|c|c|c|l}
\toprule
\textbf{Source} &
\textbf{Task} &
\textbf{Total Samples} &
\textbf{Train Samples} &
\textbf{Test Samples} &
\textbf{Metric} \\
\midrule

\multirow{4}{*}{TSQA}
& Forecasting
& 42,557
& 34,046
& 8,511
& MSE \\

& Imputation
& 38,657
& 30,926
& 7,731
& MSE \\

& Anomaly Detection
& 37,000
& 29,600
& 7,400
& Accuracy \\

& Classification
& 37,000
& 29,600
& 7,400
& Accuracy \\

\midrule

TS Instruct
& Classification
& 2,945
& 2,356
& 589
& Accuracy \\

\midrule

\textbf{Total}
& --
& \textbf{158,159}
& \textbf{126,528}
& \textbf{31,631}
& -- \\

\bottomrule
\end{tabular}
}
\caption{\small Detailed statistics of the Stage 2 task-solving datasets. }
\label{tab:stage2_dataset}
\end{table*}

\textbf{Stage 3: High-level Reasoning.}\quad
Table~\ref{tab:stage3_dataset} presents the composition of the reasoning data. The training set is collected from multiple sources, including TSQA, ChatTS, TSCognition, and TS Instruct. Two independent evaluation sets are used to assess both QA formats and five reasoning dimensions: Decoding, Grounding, Inferring, Extrapolating, and Acting.

\begin{table*}[h]
\centering
\small
\resizebox{0.9\textwidth}{!}{
\begin{tabular}{l|l|c|l}
\toprule
\textbf{Split} &
\textbf{Source} &
\textbf{\# Samples} &
\textbf{Metric} \\
\midrule

\multirow{4}{*}{Training}
& TSQA Reasoning QA
& 26,163
& -- \\

& ChatTS TSEvol + Instruct Follow
& 29,320
& -- \\

& TSCognition
& 32,869
& -- \\

& TS Instruct
& 12,852
& -- \\

\midrule

\multirow{3}{*}{Evaluation A}
& Binary QA
& \multirow{3}{*}{4,547}
& Accuracy \\

& Multiple-Choice QA
&
& Accuracy \\

& Open-ended QA
&
& ROUGE-L / BLEU \\

\midrule

\multirow{5}{*}{Evaluation B}
& Decoding
& \multirow{5}{*}{6,163}
& Accuracy \\

& Grounding
&
& Accuracy \\

& Inferring
&
& Accuracy \\

& Extrapolating
&
& Accuracy \\

& Acting
&
& Accuracy \\

\bottomrule
\end{tabular}
}
\caption{\small Detailed statistics of the Stage 3 high-level reasoning datasets.}
\label{tab:stage3_dataset}
\end{table*}

\clearpage
\section{Baseline}
\label{app:Baseline}
\textbf{ChatTS~\citep{xie2025chatts}.}\quad
ChatTS is a time series understanding framework that enhances LLMs with instruction-based temporal data covering fundamental patterns such as trends, seasonality, anomalies, and local variations. It converts time series into representations compatible with language models and performs instruction tuning for temporal question answering. We use ChatTS as a representative baseline for evaluating whether models trained on temporal understanding data can generalize to more comprehensive perception and reasoning tasks.

\textbf{TimeOmni-1~\citep{guan2025timeomni}.}\quad
TimeOmni-1 is a unified TS-LLM designed to support multiple time series understanding and reasoning tasks within a single framework. It integrates temporal information with language instructions and is trained on diverse time series tasks to improve cross-task generalization. We include TimeOmni-1 as a representative general-purpose TS-LLM baseline for comparing comprehensive temporal perception and reasoning capabilities.

\textbf{ITFormer~\citep{wang2025itformer}.}\quad
ITFormer is an end-to-end multimodal framework for time series question answering that jointly models temporal representations and language generation. It directly integrates time series features with the language model to support temporal understanding and reasoning over multimodal inputs. We use ITFormer as a representative alignment-based TS-LLM baseline for evaluating high-level temporal reasoning and out-of-domain generalization.

\textbf{PATRA~\citep{lu2026patra}.}\quad
PATRA is a recent reasoning-oriented TS-LLM developed for complex temporal analysis and question answering. It emphasizes the integration of time series representations with language reasoning to improve performance on diverse temporal reasoning tasks. We include PATRA as a strong recent baseline for comparing the reasoning and generalization capabilities of our progressively trained TS-LLMs.

\textbf{Qwen3-4B~\citep{yang2025qwen3}.}\quad
Qwen3-4B is the base large language model used in our framework. We adopt it as a non-specialized LLM baseline by directly applying the same input format without progressive time series capability training. This baseline is used to evaluate the temporal understanding and reasoning ability provided by the pretrained language model itself, and to measure the gains introduced by our stage-wise Perceive $\rightarrow$ Solve $\rightarrow$ Reason training strategy.
\section{Dataset}
\label{app:Dataset}
\textbf{ChatTS~\citep{xie2025chatts}.}\quad
ChatTS constructs fully synthetic time series--text data for temporal understanding and reasoning. Its training data consist of five subsets: UTS, MTS-Shape, MTS-Local, TSEvol, and Instruct Follow. UTS contains 35K samples for basic univariate temporal attributes, including global and local patterns; MTS-Shape contains 35K multivariate samples focusing on global trend correlations; and MTS-Local contains 35K samples emphasizing correlated local fluctuations across variables. In addition, TSEvol provides 24,270 diverse question--answer pairs for temporal reasoning, while Instruct Follow contains 5,050 samples for instruction-following ability. The generated time series cover lengths from 64 to 1024 and include numerical supervision such as extrema, segmented averages, local fluctuation positions and amplitudes, seasonality, trends, and point-wise values. In our framework, the three alignment subsets are mainly used for Stage 1 temporal perception, while TSEvol and Instruct Follow are incorporated into Stage 3 reasoning training.

\textbf{TSQA~\citep{kong2025time}.}\quad
TSQA is a large-scale time series QA dataset introduced with Time-MQA, containing approximately 200K QA pairs across 12 real-world application domains, including healthcare, environment, energy, finance, transportation, IoT, AIOps, and human activities. It covers five task categories: forecasting, imputation, anomaly detection, classification, and open-ended reasoning. Specifically, it contains 42,557 forecasting, 38,657 imputation, 37,000 anomaly detection, 37,000 classification, and 37,629 open-ended QA samples. The reasoning subset further includes true/false, multiple-choice, and open-ended questions. In our framework, the numerical task subsets are mainly used in Stage 2, while the reasoning QA subset is incorporated into Stage 3.

\textbf{TS Instruct~\citep{quinlan2026chat}.}\quad
TS Instruct is a multimodal instruction-tuning dataset introduced with Chat-TS, pairing diverse time series with natural-language instructions and responses. The training set contains 18,306 samples covering four categories: 9,726 reasoning samples, 3,126 decision-making samples, 2,945 classification samples, and 2,509 mathematical reasoning samples. The dataset is generated using time series properties, visual representations, task descriptors, and synthetic conversations to support joint reasoning over temporal and textual information. In our framework, the classification subset is used in Stage 2, while the remaining reasoning-oriented samples are incorporated into Stage 3.

\begin{figure}[h]
    \centering
    \includegraphics[width=1\linewidth]{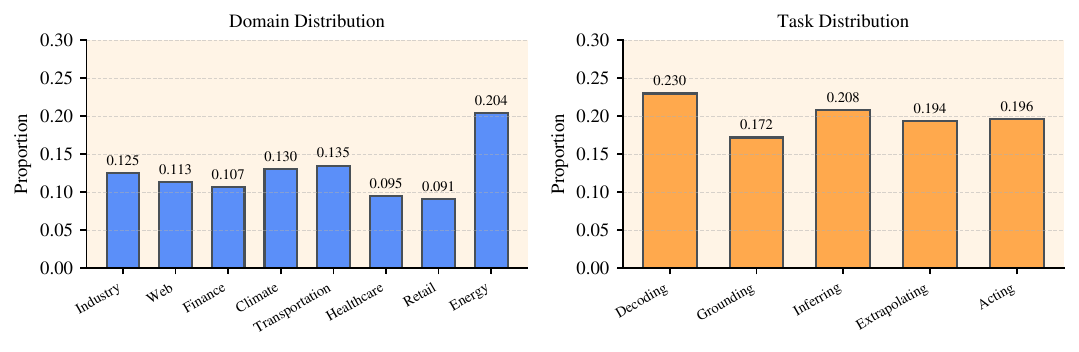}
    \caption{Distribution of TSCognition samples across eight domains and five hierarchical tasks.}
    \label{fig:dataset_distribution}
\end{figure}

\textbf{TSCognition~\citep{qiu2026recognition}.}\quad
TSCognition is constructed from 15 public real-world datasets spanning eight domains, including Energy (Buildings-900K\footnote{\scriptsize\url{https://data.openei.org/submissions/5859}} and UMass Smart*\footnote{\scriptsize\url{https://traces.cs.umass.edu/docs/traces/smartstar/}}), Retail (Favorita Sales\footnote{\scriptsize\url{https://www.kaggle.com/competitions/favorita-grocery-sales-forecasting/data}} and Dominick\footnote{\scriptsize\url{https://www.chicagobooth.edu/research/kilts/datasets/dominicks}}), Healthcare (Project Tycho\footnote{\scriptsize\url{https://www.tycho.pitt.edu/}} and JHU CSSE COVID-19\footnote{\scriptsize\url{https://github.com/CSSEGISandData/COVID-19}}), Transportation (LargeST\footnote{\scriptsize\url{https://github.com/liuxu77/LargeST}} and METR-LA\footnote{\scriptsize\url{https://github.com/liyaguang/DCRNN}}), Climate (SubseasonalClimateUSA\footnote{\scriptsize\url{https://github.com/microsoft/subseasonal_data}} and Oikolab Weather\footnote{\scriptsize\url{https://docs.oikolab.com/datasets/}}), Finance (Bitcoin\footnote{\scriptsize\url{https://zenodo.org/records/5122101}}), Web (Web Traffic\footnote{\scriptsize\url{https://www.kaggle.com/competitions/web-traffic-time-series-forecasting/data}}), and Industry (WADI\footnote{\scriptsize\url{https://www.sutd.edu.sg/itrust/itrust-labs/datasets/dataset-characteristics/wadi/}}, C-MAPSS\footnote{\scriptsize\url{https://data.nasa.gov/dataset/c-mapss-aircraft-engine-simulator-data}}, and Paderborn\footnote{\scriptsize\url{https://mb.uni-paderborn.de/en/kat/research/bearing-datacenter}}). It organizes time series cognition into five hierarchical tasks: \textit{Decoding}, \textit{Grounding}, \textit{Inferring}, \textit{Extrapolating}, and \textit{Acting}. Decoding focuses on extracting fundamental temporal patterns, Grounding aligns temporal signals with contextual information, Inferring models dependencies and relational structures, Extrapolating extends learned patterns to future dynamics, and Acting integrates preceding reasoning results to support decision making. The sample distribution across domains and tasks is shown in Figure~\ref{fig:dataset_distribution}, while detailed task-level statistics, including variable numbers, time series lengths, and answer choices, are summarized in Table~\ref{tab:task_level_statistics}. In our framework, the TSCognition training split is incorporated into Stage 3 to strengthen high-level temporal reasoning capability.

\begin{table}[h]
\centering

\renewcommand{\arraystretch}{1.05}
\resizebox{0.75\linewidth}{!}{
\begin{tabular}{l|ccc|ccc|ccc}
\toprule
\multirow{2}{*}{\textbf{Tasks}} 
& \multicolumn{3}{c|}{\textbf{ Variable Numbers}} 
& \multicolumn{3}{c|}{\textbf{Time Series Length}} 
& \multicolumn{3}{c}{\textbf{ Choices}} \\
\cmidrule(lr){2-4} \cmidrule(lr){5-7} \cmidrule(lr){8-10}
& \textbf{Min} & \textbf{Max} & \textbf{Avg.}
& \textbf{Min} & \textbf{Max} & \textbf{Avg.}
& \textbf{Min} & \textbf{Max} & \textbf{Avg.} \\
\midrule
Decoding Task
& 4 & 5 & 4.7 
& 437 & 982  & 742 
& 4 & 5 & 4.7 \\

Grounding Task
& 3 & 6 & 4.4 
& 656 & 1018 & 871 
& 4 & 5 & 4.3 \\

Inferring Task
& 3 & 6 & 5.1 
& 882 & 1059 & 912 
& 4 & 5 & 4.6 \\

Extrapolating Task
& 3 & 6 & 5.2 
& 601 & 1052 & 846 
& 4 & 5 & 4.5 \\

Acting Task
& 4 & 6 & 5.5 
& 468 & 1047 & 795 
& 4 & 5 & 4.2 \\
\midrule
Overall 
& 3 & 6 & 4.9 
& 437 & 1059 & 785 
& 4 & 5 & 4.5 \\
\bottomrule
\end{tabular}
}
\caption{Task-level statistics of time series in TSCognition.}
\label{tab:task_level_statistics}
\end{table}

\begin{wraptable}{r}{0.4\linewidth}
\centering
\vspace{-0.1cm}
\small
\resizebox{0.8\linewidth}{!}{
\begin{tabular}{l|c}
\toprule
\textbf{Category} &
\textbf{Samples} \\
\midrule

Anomaly Detection & 108 \\
Causality & 72 \\
Noise Understanding & 84 \\
Pattern Recognition & 362 \\
Similarity Analysis & 120 \\

\midrule
\textbf{Total} & \textbf{746} \\

\bottomrule
\end{tabular}
}
\vspace{-0.15cm}
\caption{\small Statistics of TimeSeriesExam.}
\label{tab:timeseriesexam_statistics}
\vspace{-0.4cm}
\end{wraptable}
\textbf{TimeSeriesExam~\citep{cai2024timeseriesexam}.}\quad
TimeSeriesExam is a configurable multiple-choice benchmark designed to evaluate fundamental time series understanding in LLMs. It covers five categories: \textit{pattern recognition}, \textit{noise understanding}, \textit{similarity analysis}, \textit{anomaly detection}, and \textit{causality analysis}. The benchmark is procedurally generated from 104 expert-designed question templates and contains more than 700 questions. Its synthetic time series are constructed from diverse baseline signal components, including linear, nonlinear, and cyclic patterns, allowing controlled evaluation of different temporal properties. The detailed category statistics  are summarized in Table~\ref{tab:timeseriesexam_statistics}.

\textbf{CaTS-Bench~\citep{zhou2026cats}.}\quad
CaTS-Bench is a real-world benchmark for context-aware time series captioning and reasoning across 11 diverse domains. Each sample combines a numeric time series segment with contextual metadata, a line-chart visualization, and a natural-language caption. In addition to caption generation, CaTS-Bench provides diagnostic multiple-choice tasks for evaluating fine-grained temporal understanding, including caption retrieval, time series retrieval, amplitude comparison, and mean comparison in Table~\ref{tab:catsbench_statistics}. In our experiments, we use these diagnostic tasks to evaluate whether the learned temporal perception capability generalizes to unseen real-world data and cross-modal reasoning settings. 

\begin{table}[h]
\centering
\small
\resizebox{0.3\linewidth}{!}{
\begin{tabular}{l|c}
\toprule
\textbf{Task} & \textbf{Samples} \\
\midrule
Caption Retrieval & 100 \\
Time Series Retrieval & 100 \\
Amplitude Comparison & 40 \\
Mean Comparison & 40 \\
\midrule
\textbf{Total} & \textbf{280} \\
\bottomrule
\end{tabular}
}
\caption{\small Statistics of CaTS-Bench used in our evaluation.}
\label{tab:catsbench_statistics}
\end{table}

\textbf{MTBench~\citep{chen2025mtbench}.}\quad
MTBench is a multimodal time series benchmark for evaluating temporal reasoning over numerical signals and textual context in the weather and finance domains. It includes weather question answering and prediction, as well as short- and long-horizon financial trend prediction and forecasting tasks. We use MTBench exclusively for out-of-domain evaluation to assess model generalization to unseen domains and reasoning settings. The evaluation tasks used in our experiments are summarized in Table~\ref{tab:mtbench_statistics}.

\begin{table}[h]
\centering
\small
\resizebox{0.4\linewidth}{!}{
\begin{tabular}{l|l}
\toprule
\textbf{Domain} & \textbf{Evaluation Tasks} \\
\midrule
Weather
& MCQA / Prediction \\

Finance (7 days)
& 3-way / 5-way / Forecasting \\

Finance (30 days)
& 3-way / 5-way / Forecasting \\
\bottomrule
\end{tabular}
}
\caption{\small Evaluation settings of MTBench.}
\label{tab:mtbench_statistics}
\end{table}

\section{Detailed Results}
\label{app:Results}
\textbf{Perception Results.}\quad
Table~\ref{tab:cate_num_comparison_detailed} reports the detailed categorical and numerical perception results across different methods. It compares performance on trend, seasonality, noise, local patterns, correlation, and clustering, corresponding to the perception analysis shown in Figure~\ref{fig:cate_num_comparison_four_models}.

\begin{table*}[h]
\centering
\small
\resizebox{0.98\textwidth}{!}{
\begin{tabular}{l|cc|cc|cc|cc|c|c}
\toprule
\multirow{2}{*}{\textbf{Method}} &
\multicolumn{2}{c|}{\textbf{Trend}} &
\multicolumn{2}{c|}{\textbf{Season}} &
\multicolumn{2}{c|}{\textbf{Noise}} &
\multicolumn{2}{c|}{\textbf{Local}} &
\textbf{Corr.} &
\textbf{Clus.} \\
\cmidrule(lr){2-3}
\cmidrule(lr){4-5}
\cmidrule(lr){6-7}
\cmidrule(lr){8-9}
&
\textbf{Cate.} & \textbf{Num.} &
\textbf{Cate.} & \textbf{Num.} &
\textbf{Cate.} & \textbf{Num.} &
\textbf{Cate.} & \textbf{Num.} &
\textbf{Cate.} &
\textbf{Cate.} \\
\midrule

Pre-align
& 0.937 & 0.865
& 0.911 & 0.872
& 0.857 & 0.534
& 0.799 & 0.700
& 0.831
& 0.783 \\

GPT-4o (Text)
& 0.664 & 0.839
& 0.751 & 0.546
& 0.654 & 0.287
& 0.237 & 0.165
& 0.361
& 0.294 \\

GPT-4o (Vision)
& 0.691 & 0.423
& 0.907 & 0.582
& 0.659 & 0.361
& 0.415 & 0.302
& 0.394
& 0.411 \\

Post-align
& 0.922 & 0.880
& 0.931 & 0.834
& 0.900 & 0.479
& 0.824 & 0.733
& 0.771
& 0.804 \\

\bottomrule
\end{tabular}
}
\caption{\small Detailed categorical and numerical perception results.}
\label{tab:cate_num_comparison_detailed}
\end{table*}

\textbf{Reasoning Scaling Results.}\quad
Table~\ref{tab:reasoning_data_scaling_detailed} reports the detailed reasoning accuracy under different proportions of Stage 3 training data, corresponding to Figure~\ref{fig:reasoning_data_scaling}. For both Pre-align and Post-align architectures, progressive Stage (1$\rightarrow$2$\rightarrow$3) training consistently outperforms Stage 3 Only and Mixed Training, with a larger advantage under limited reasoning supervision. The Directly setting shows weaker overall performance, further indicating that explicit temporal representation learning benefits progressive capability formation.

\begin{table*}[h]
\centering
\small
\resizebox{0.8\textwidth}{!}{
\begin{tabular}{l|l|ccccc}
\toprule
\textbf{Model} &
\textbf{Training Setting} &
\textbf{15\%} &
\textbf{25\%} &
\textbf{50\%} &
\textbf{75\%} &
\textbf{100\%} \\
\midrule

\multirow{3}{*}{Pre-align}
& Stage 3 Only
& 0.414 & 0.464 & 0.527 & 0.586 & 0.619 \\

& Mixed Training
& 0.479 & 0.520 & 0.583 & 0.673 & 0.689 \\

& Stage (1$\rightarrow$2$\rightarrow$3)
& 0.528 & 0.574 & 0.598 & 0.664 & 0.710 \\

\midrule

\multirow{3}{*}{Post-align}
& Stage 3 Only
& 0.403 & 0.453 & 0.524 & 0.585 & 0.617 \\

& Mixed Training
& 0.461 & 0.502 & 0.570 & 0.664 & 0.678 \\

& Stage (1$\rightarrow$2$\rightarrow$3)
& 0.520 & 0.561 & 0.614 & 0.680 & 0.701 \\

\midrule

\multirow{6}{*}{Directly}
& Stage 3 Only
& 0.402 & 0.436 & 0.478 & 0.517 & 0.546 \\

& Mixed Training
& 0.439 & 0.497 & 0.530 & 0.588 & 0.608 \\

& Stage (1$\rightarrow$2$\rightarrow$3)
& 0.451 & 0.490 & 0.542 & 0.579 & 0.614 \\

& Stage (1$\rightarrow$3)
& 0.429 & 0.482 & 0.519 & 0.572 & 0.599 \\

& Stage (2$\rightarrow$3)
& 0.447 & 0.469 & 0.535 & 0.563 & 0.604 \\

& No Training
& 0.351 & 0.351 & 0.351 & 0.351 & 0.351 \\

\bottomrule
\end{tabular}
}
\caption{\small Detailed reasoning accuracy under different Stage 3 training data scales.}
\label{tab:reasoning_data_scaling_detailed}
\end{table*}

\textbf{Stage-wise Saturation Results.}\quad
Table~\ref{tab:stage_saturation_detailed} provides the detailed results corresponding to Table~\ref{tab:stage_saturation}. Stage 1 and Stage (1$\rightarrow$2) reach local saturation at approximately 70\% of the corresponding training data, achieving 98.3\% and 98.8\% of their full-stage performance, respectively. Stage (1$\rightarrow$2$\rightarrow$3) reaches a similar saturation region at 85\%, with 97.2\% of its full-stage performance. These results further show that local convergence can occur before the complete progressive pathway reaches its best performance.
\begin{table*}[h]
\centering
\small
\resizebox{0.92\textwidth}{!}{
\begin{tabular}{l|c|c|c|c}
\toprule
\textbf{Training Setting} &
\textbf{Data Scale} &
\textbf{Evaluation Performance} &
\textbf{Relative Perf.} &
\textbf{Full Progressive} \\
\midrule

\multirow{5}{*}{Stage 1}
& 40\%  & 0.613 & -- & \multirow{5}{*}{0.743} \\
& 55\%  & 0.662 & -- & \\
& \textbf{70\%}  & \textbf{0.738} & \textbf{98.3\%} & \\
& 85\%  & 0.745 & -- & \\
& 100\% & 0.751 & 100\% & \\

\midrule

\multirow{5}{*}{Stage (1$\rightarrow$2)}
& 40\%  & 0.542 & -- & \multirow{5}{*}{0.684} \\
& 55\%  & 0.603 & -- & \\
& \textbf{70\%}  & \textbf{0.644} & \textbf{98.8\%} & \\
& 85\%  & 0.649 & -- & \\
& 100\% & 0.652 & 100\% & \\

\midrule

\multirow{5}{*}{Stage (1$\rightarrow$2$\rightarrow$3)}
& 40\%  & 0.537 & -- & \multirow{5}{*}{0.710} \\
& 55\%  & 0.569 & -- & \\
& 70\%  & 0.616 & -- & \\
& \textbf{85\%}  & \textbf{0.663} & \textbf{97.2\%} & \\
& 100\% & 0.682 & 100\% & \\

\bottomrule
\end{tabular}
}
\caption{\small Detailed stage-wise performance under increasing training data scales. For Stage (1$\rightarrow$2), Stage 1 is fixed at its 70\% saturation point; for Stage (1$\rightarrow$2$\rightarrow$3), Stages 1 and 2 are both fixed at 70\%. \textbf{Bold} denotes the identified local saturation points.}
\label{tab:stage_saturation_detailed}
\end{table*}
\section{Supplementary Experiments}
\label{Supplementary}
To examine whether the observed progressive capability formation is specific to the Qwen3-4B backbone, we further conduct supplementary experiments using Qwen3-1.7B. We follow the same three-stage training paradigm and evaluation protocol as the main experiments. The results across perception, task solving, and high-level reasoning show similar trends, indicating that the progressive capability relationships remain consistent under a smaller LLM backbone.

\textbf{Stage 1: Fundamental Attribute Perception.}\quad
Table~\ref{tab:qwen17_stage1} reports the Stage 1 perception results of Pre-align and Post-align. They show strong performance across categorical and numerical temporal attributes, including trend, seasonality, noise, local patterns, correlation, and clustering.

\begin{table*}[h]
\centering
\small
\resizebox{0.96\textwidth}{!}{
\begin{tabular}{l|cc|cc|cc|cc|cc}
\toprule
\multirow{2}{*}{\textbf{Model}} &
\multicolumn{2}{c|}{\textbf{Trend}} &
\multicolumn{2}{c|}{\textbf{Season}} &
\multicolumn{2}{c|}{\textbf{Noise}} &
\multicolumn{2}{c|}{\textbf{Local}} &
\multicolumn{2}{c}{\textbf{Multivariate}} \\
\cmidrule(lr){2-3}
\cmidrule(lr){4-5}
\cmidrule(lr){6-7}
\cmidrule(lr){8-9}
\cmidrule(lr){10-11}
&
\textbf{Cate.} & \textbf{Num.} &
\textbf{Cate.} & \textbf{Num.} &
\textbf{Cate.} & \textbf{Num.} &
\textbf{Cate.} & \textbf{Num.} &
\textbf{Corr.} & \textbf{Clus.} \\
\midrule

Pre-align
& 0.894 & 0.843
& 0.859 & 0.843
& 0.822 & 0.558
& 0.786 & 0.727
& 0.788 & 0.761 \\

Post-align
& 0.873 & 0.846
& 0.892 & 0.834
& 0.861 & 0.506
& 0.752 & 0.709
& 0.755 & 0.747 \\

\bottomrule
\end{tabular}
}
\caption{\small Stage 1 perception results on Qwen3-1.7B.}
\label{tab:qwen17_stage1}
\end{table*}

\textbf{Stage 2: Temporal Task Solving.}\quad
Table~\ref{tab:qwen17_stage2} presents the Stage 2 results. Stage 1-only training already improves all four downstream tasks over the non-fine-tuned Qwen3-1.7B baseline. Further task-specific supervision provides substantial gains, while Stage (1$\rightarrow$2) achieves competitive or better performance on most metrics.

\begin{table*}[h]
\centering
\small
\resizebox{0.90\textwidth}{!}{
\begin{tabular}{l|l|cccc}
\toprule
\textbf{Model} &
\textbf{Training Setting} &
\textbf{Fore. MSE $\downarrow$} &
\textbf{Impu. MSE $\downarrow$} &
\textbf{Ano. Det. Acc. $\uparrow$} &
\textbf{Class. Acc. $\uparrow$} \\
\midrule

Qwen3-1.7B
& No Fine-tuning
& 1.884 & 0.0312 & 0.303 & 0.234 \\

\midrule

\multirow{3}{*}{Pre-align}
& Stage 1 Only
& 1.813 & 0.0218 & 0.548 & 0.421 \\

& Stage 2 Only
& \textbf{1.412} & 0.0151 & 0.651 & 0.542 \\

& Stage (1$\rightarrow$2)
& 1.438 & \textbf{0.0144} & \textbf{0.684} & \textbf{0.576} \\

\midrule

\multirow{3}{*}{Post-align}
& Stage 1 Only
& 1.782 & 0.0235 & 0.574 & 0.408 \\

& Stage 2 Only
& 1.469 & 0.0162 & \textbf{0.648} & 0.507 \\

& Stage (1$\rightarrow$2)
& \textbf{1.391} & \textbf{0.0150} & 0.626 & \textbf{0.559} \\

\bottomrule
\end{tabular}
}
\caption{\small Stage 2 task-solving results on Qwen3-1.7B.}
\label{tab:qwen17_stage2}
\end{table*}

\textbf{Stage 3: High-level Reasoning.}\quad
Table~\ref{tab:qwen17_stage3} reports the reasoning results on the Stage 3 evaluation set B. Compared with Stage 3 Only, incorporating preceding perception or task-solving stages consistently improves reasoning performance. The complete Stage (1$\rightarrow$2$\rightarrow$3) pathway achieves the strongest overall results across the five reasoning dimensions.

\begin{table*}[h]
\centering
\small
\resizebox{0.92\textwidth}{!}{
\begin{tabular}{l|l|ccccc}
\toprule
\textbf{Model} &
\textbf{Training Setting} &
\textbf{Decoding} &
\textbf{Grounding} &
\textbf{Inferring} &
\textbf{Extrapolating} &
\textbf{Acting} \\
\midrule

\multirow{4}{*}{Pre-align}
& Stage 3 Only
& 65.8 & 63.7 & 54.2 & 53.6 & 46.6 \\

& Stage (1$\rightarrow$3)
& 73.8 & \textbf{76.9} & 60.7 & 58.9 & 49.4 \\

& Stage (2$\rightarrow$3)
& 74.9 & 69.5 & \textbf{65.6} & 58.1 & 52.1 \\

& Stage (1$\rightarrow$2$\rightarrow$3)
& \textbf{78.1} & 74.2 & 63.1 & \textbf{65.2} & \textbf{55.6} \\

\midrule

\multirow{4}{*}{Post-align}
& Stage 3 Only
& 62.9 & 62.8 & 53.1 & 55.4 & 47.7 \\

& Stage (1$\rightarrow$3)
& 72.5 & 70.3 & 59.4 & 61.5 & 50.9 \\

& Stage (2$\rightarrow$3)
& 70.4 & 69.8 & \textbf{64.2} & 63.4 & \textbf{57.3} \\

& Stage (1$\rightarrow$2$\rightarrow$3)
& \textbf{75.4} & \textbf{74.3} & 61.6 & \textbf{67.7} & 53.2 \\

\bottomrule
\end{tabular}
}
\caption{\small Stage 3 reasoning results on Evaluation Set B using Qwen3-1.7B.}
\label{tab:qwen17_stage3}
\end{table*}

Overall, the Qwen3-1.7B exhibit trends consistent with the main Qwen3-4B results. Fundamental perception provides transferable temporal knowledge for downstream task solving, while the combination of perception and task-solving supervision further supports high-level reasoning. These results suggest that the observed capability formation is not restricted to a specific model scale.

% \clearpage
\section{Reproducibility}
\label{app:Reproducibility}

We adopt unified implementation, training, and evaluation settings whenever possible to ensure fair and reproducible comparisons in Talbe~\ref{tab:implementation_details}. Unless otherwise specified, all experiments are conducted on two NVIDIA A800 GPUs using PyTorch, HuggingFace Transformers, DeepSpeed, and Accelerate. We use Qwen3-4B as the primary LLM backbone and additionally conduct supplementary experiments on Qwen3-1.7B. The same data splits, stage definitions, evaluation protocols, and prompt formats are maintained within each group of comparative experiments. We fix the random seed to 2026 and use bfloat16 mixed-precision training for all LLM-based experiments.

\begin{table}[h]
\centering
\small
\renewcommand{\arraystretch}{1.05}
\resizebox{0.72\linewidth}{!}{
\begin{tabular}{l|c}
\toprule
\textbf{Configuration} & \textbf{Setting} \\
\midrule
Primary backbone & Qwen3-4B \\
Additional backbone & Qwen3-1.7B \\
GPUs & 2 $\times$ NVIDIA A800 \\
Framework & PyTorch + HuggingFace Transformers \\
Distributed training & DeepSpeed + Accelerate \\
Precision & bfloat16 \\
Optimizer & AdamW \\
Learning rate & $2 \times 10^{-5}$ \\
Weight decay & $1 \times 10^{-2}$ \\
Batch size per GPU & 4 \\
Gradient accumulation steps & 4 \\
Effective batch size & 32 \\
Gradient clipping & 1.0 \\
Learning rate scheduler & Cosine decay \\
Warmup ratio & 0.0 \\
Random seed & 2026 \\
\bottomrule
\end{tabular}
}
\caption{\small Implementation and training configurations used in our experiments.}
\label{tab:implementation_details}
\end{table}

All capability stages are trained using supervised fine-tuning with the same optimization configuration unless otherwise specified. For the Pre-align architecture, the Temporal Encoder and alignment module are optimized in Stage 1, while the Temporal Encoder is frozen in subsequent stages and the remaining alignment components are updated. For the Post-align architecture, the Temporal Encoder and LLM backbone are jointly optimized. For the textualization setting, the numerical time series are directly converted into textual sequences and only the LLM is fine-tuned. Across different capability-development paths, we keep the model architecture, data split, initialization, and evaluation procedure unchanged so that performance differences mainly reflect the training stage and supervision strategy. This ensures a controlled comparison across different training paradigms.

For evaluation, Stage 1 separately measures categorical and numerical temporal perception, Stage 2 evaluates forecasting, imputation, anomaly detection, and classification, and Stage 3 evaluates high-level temporal reasoning. Forecasting and imputation use MSE, anomaly detection and classification use accuracy, while the reasoning dimensions are evaluated using multiple-choice accuracy when applicable. All reported comparative results are obtained under the same evaluation protocol within each experiment. This provides consistent evaluation criteria across all capability stages.

% \clearpage
\section{Case Study and Failure Analysis}
\label{app:Case Study and Failure Analysis}
\begin{figure}[h]
\begin{tcolorbox}[
    title=\textbf{Example of the Decoding task},
    colback=mycyan,
    colframe=mycyan1,
    boxrule=0.6pt,
    arc=2pt,
    left=5pt,
    right=5pt,
    top=5pt,
    bottom=5pt
]
\textbf{Instruction:} You are an expert in time series analysis and temporal reasoning. A temporal token representing the input time series is provided to the model. Please answer the multiple-choice question based on the temporal evidence, context, and candidate answers. Only one answer is correct.

\textbf{Variable Information:} 
$D$: Load Demand; 
$G$: Thermal Generation; 
$R$: Renewable Output; 
$B$: Battery Power; 
$P$: Electricity Price; 
$F$: Frequency Deviation.

\vspace{2mm}
\begin{center}
    \includegraphics[width=0.92\linewidth]{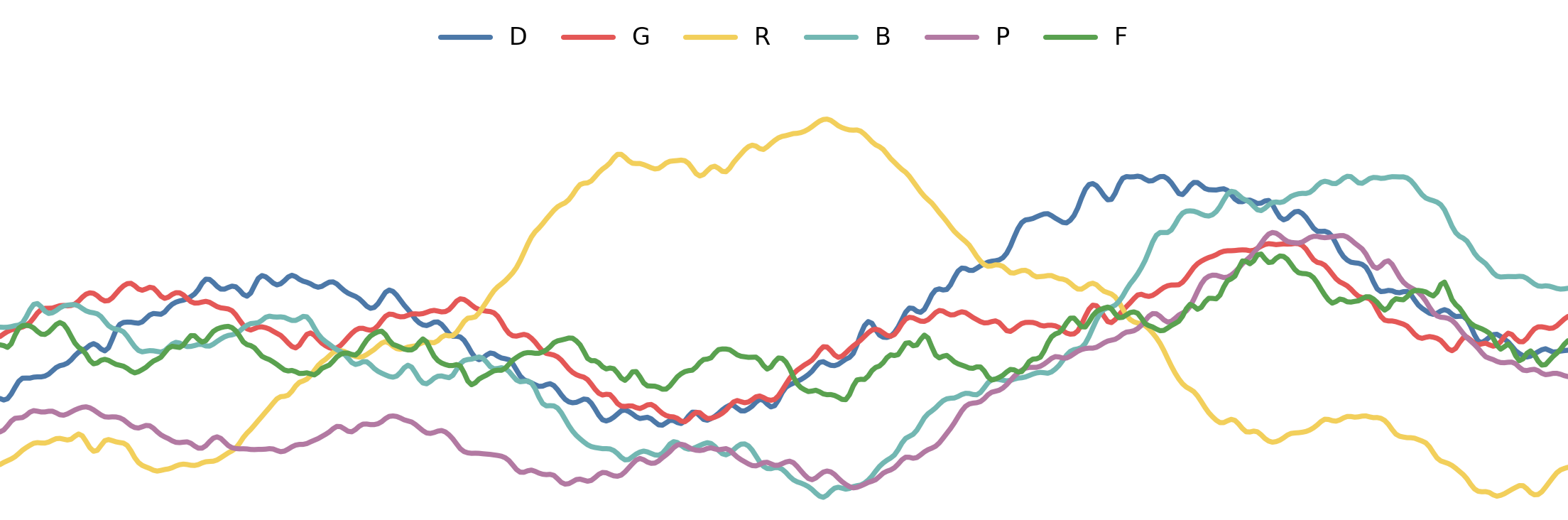}
\end{center}
\vspace{1mm}

\textbf{Context:} 
The data describe the operation of a regional power system. 
The operator monitors load demand, thermal generation, renewable output, battery behavior, electricity price, and frequency deviation to evaluate supply-demand balance and dispatch decisions. 
Battery power reflects storage operation, while electricity price and frequency deviation may indicate system stress. 
The actual temporal state should be inferred from the provided time series representation.

\textbf{Question:} 
Which statement best characterizes the joint temporal behavior of renewable output and battery power?

\textbf{Candidate Answers:} \\
A. Renewable output stays low throughout the period, while the battery remains inactive with no clear charging or discharging phase. \\
B. Renewable output reaches its highest level only during the final high-demand period, while the battery mainly charges at the same time. \\
C. Renewable output and battery power increase monotonically together, indicating that storage output always follows renewable generation. \\
D. Renewable output first increases to a mid-period peak and then declines, while the battery tends to charge during the renewable-rich phase and discharge later. \\
E. Renewable output fluctuates randomly without a recognizable phase, while battery behavior shows no relation to the system state.

\textbf{Output:} Please output the correct candidate answer.

\vspace{1mm}

\textbf{Answer:} \textcolor{answergreen}{\textit{D. Renewable output forms a clear mid-period peak
  before decreasing. Battery power shifts from charging during the renewable-rich phase to discharging
  afterward.}}

\end{tcolorbox}
\end{figure}

\begin{figure}[h]
\begin{tcolorbox}[
    title=\textbf{Example of the Grounding task},
    colback=mycyan,
    colframe=mycyan1,
    boxrule=0.6pt,
    arc=2pt,
    left=5pt,
    right=5pt,
    top=5pt,
    bottom=5pt
]
\textbf{Instruction:} You are an expert in time series analysis and temporal reasoning. A temporal token representing the input time series is provided to the model. Please answer the multiple-choice question based on the temporal evidence, context, and candidate answers. Only one answer is correct.

\textbf{Variable Information:} 
$D$: Load Demand; 
$G$: Thermal Generation; 
$R$: Renewable Output; 
$B$: Battery Power; 
$P$: Electricity Price; 
$F$: Frequency Deviation.

\vspace{2mm}
\begin{center}
    \includegraphics[width=0.92\linewidth]{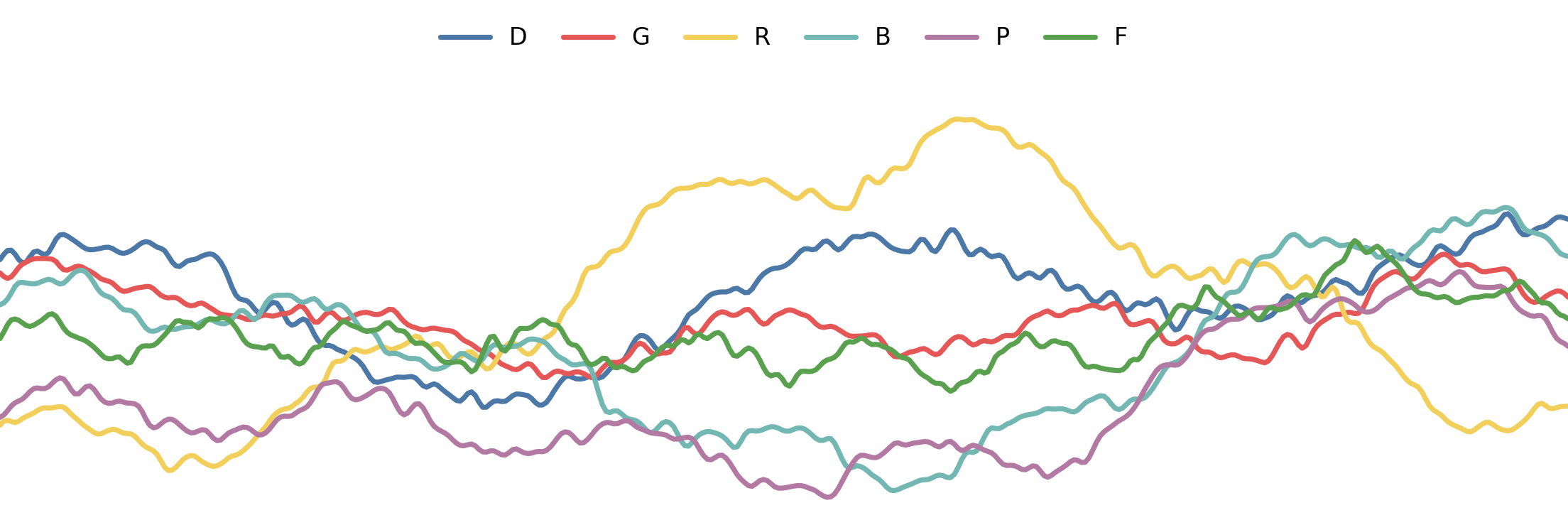}
\end{center}
\vspace{1mm}

\textbf{Context:} 
The data describe the operation of a regional power system. 
The operator monitors load demand, thermal generation, renewable output, battery behavior, electricity price, and frequency deviation to evaluate supply-demand balance and dispatch decisions. 
Battery power reflects storage operation, while electricity price and frequency deviation may indicate system stress. 
The actual temporal state should be inferred from the provided time series representation.

\textbf{Question:} 
Which interpretation best grounds the observed temporal patterns in the operational context of power dispatch?

\textbf{Candidate Answers:} \\
A. A renewable-rich period is reflected by higher renewable output together with battery charging, suggesting that surplus renewable energy is being absorbed. \\
B. A renewable-rich period is reflected by high electricity price and large frequency deviation, because these always indicate abundant renewable supply. \\
C. A peak-load period is reflected by low demand and strong battery charging, suggesting that the system is under severe supply shortage. \\
D. A stable operating period is reflected by simultaneous increases in load demand, electricity price, and frequency deviation, regardless of renewable output. \\
E. The storage behavior cannot be grounded in the power-system context because battery power is unrelated to renewable output or demand.

\textbf{Output:} Please output the correct candidate answer.

\textbf{Answer:} \textcolor{answergreen}{\textit{A. The high-renewable interval coincides with
  sustained battery charging. This pattern indicates that the battery is absorbing surplus renewable
  energy.}}
  
\end{tcolorbox}
\end{figure}

\begin{figure}[h]
\begin{tcolorbox}[
    title=\textbf{Example of the Inferring task},
    colback=mycyan,
    colframe=mycyan1,
    boxrule=0.6pt,
    arc=2pt,
    left=5pt,
    right=5pt,
    top=5pt,
    bottom=5pt
]
\textbf{Instruction:} You are an expert in time series analysis and temporal reasoning. A temporal token representing the input time series is provided to the model. Please answer the multiple-choice question based on the temporal evidence, context, and candidate answers. Only one answer is correct.

\textbf{Variable Information:} 
$D$: Load Demand; 
$G$: Thermal Generation; 
$R$: Renewable Output; 
$B$: Battery Power; 
$P$: Electricity Price; 
$F$: Frequency Deviation.

\vspace{2mm}
\begin{center}
    \includegraphics[width=0.92\linewidth]{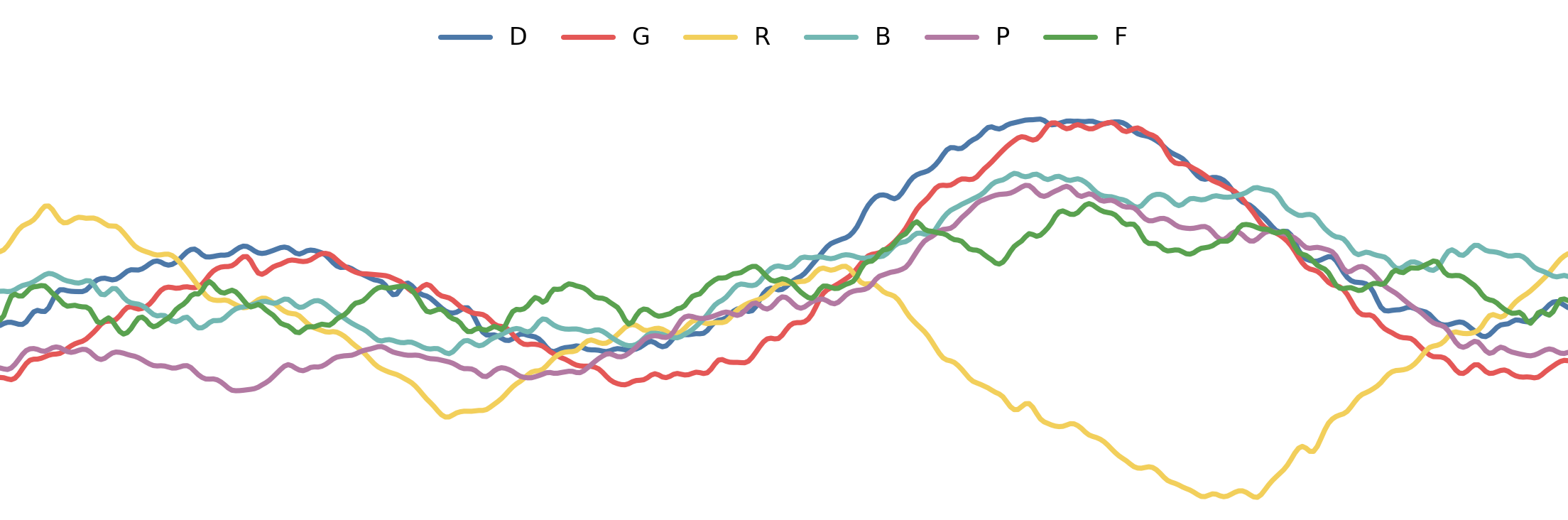}
\end{center}
\vspace{1mm}

\textbf{Context:} 
The data describe the operation of a regional power system. 
The operator monitors load demand, thermal generation, renewable output, battery behavior, electricity price, and frequency deviation to evaluate supply-demand balance and dispatch decisions. 
Battery power reflects storage operation, while electricity price and frequency deviation may indicate system stress. 
The actual temporal state should be inferred from the provided time series representation.

\textbf{Question:} 
What is the most plausible explanation for the system pressure indicated by electricity price and frequency deviation during the high-load phase?

\textbf{Candidate Answers:} \\
A. The pressure is mainly caused by decreasing demand, since lower load usually forces the system to increase electricity price and frequency deviation. \\
B. The pressure is likely caused by the coincidence of high load demand, reduced renewable support, and increased dependence on dispatchable resources. \\
C. The pressure is mainly caused by excessive renewable generation, because abundant renewable output always increases electricity price and frequency deviation. \\
D. The pressure is unrelated to supply-demand balance and is only caused by random fluctuations in the battery power curve. \\
E. The pressure disappears because battery discharge and thermal generation fully eliminate all frequency deviation.

\textbf{Output:} Please output the correct candidate answer.

\textbf{Answer:} \textcolor{answergreen}{\textit{B. Load demand rises while renewable output weakens
  during the high-pressure phase. The concurrent increases in thermal generation, electricity price, and
  frequency deviation indicate greater reliance on dispatchable resources.}}
\end{tcolorbox}
\end{figure}

\begin{figure}[h]
\begin{tcolorbox}[
    title=\textbf{Example of the Extrapolating task},
    colback=mycyan,
    colframe=mycyan1,
    boxrule=0.6pt,
    arc=2pt,
    left=5pt,
    right=5pt,
    top=5pt,
    bottom=5pt
]
\textbf{Instruction:} You are an expert in time series analysis and temporal reasoning. A temporal token representing the input time series is provided to the model. Please answer the multiple-choice question based on the temporal evidence, context, and candidate answers. Only one answer is correct.

\textbf{Variable Information:} 
$D$: Load Demand; 
$G$: Thermal Generation; 
$R$: Renewable Output; 
$B$: Battery Power; 
$P$: Electricity Price; 
$F$: Frequency Deviation.

\vspace{2mm}
\begin{center}
    \includegraphics[width=0.92\linewidth]{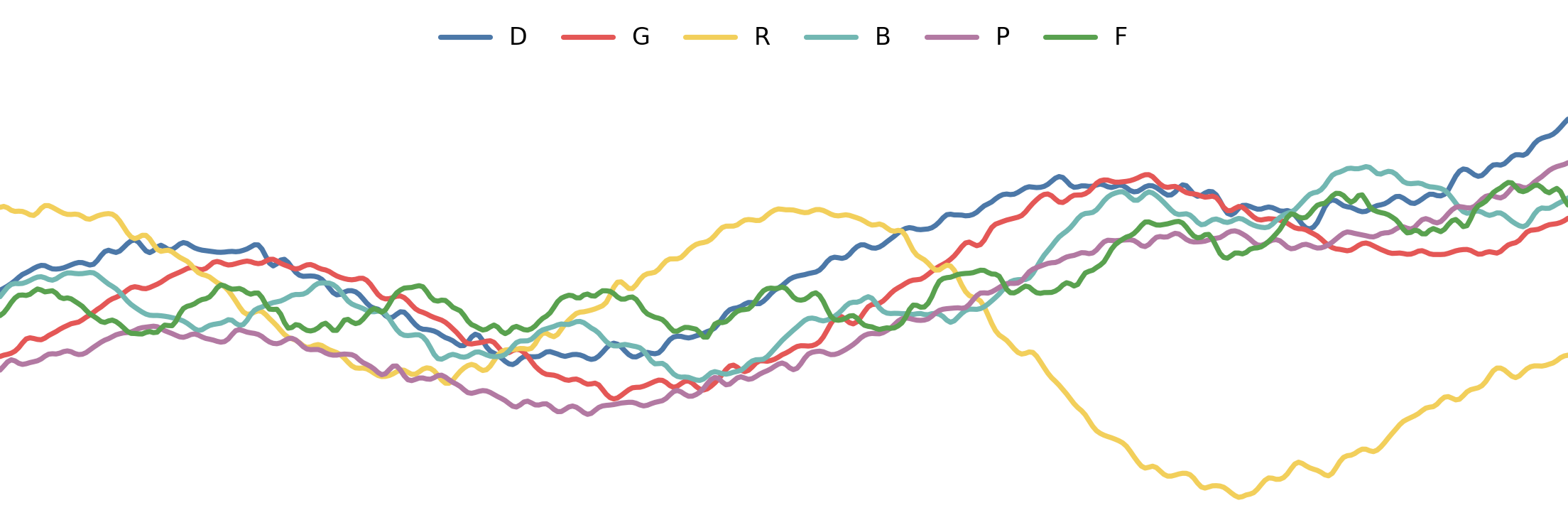}
\end{center}
\vspace{1mm}

\textbf{Context:} 
The data describe the operation of a regional power system. 
The operator monitors load demand, thermal generation, renewable output, battery behavior, electricity price, and frequency deviation to evaluate supply-demand balance and dispatch decisions. 
Battery power reflects storage operation, while electricity price and frequency deviation may indicate system stress. 
The actual temporal state should be inferred from the provided time series representation.

\textbf{Question:} 
Assume that the current high-load phase continues for the next interval and renewable output does not recover. Which future evolution is most consistent with the observed temporal evidence?

\textbf{Candidate Answers:} \\
A. The system will remain risk-free even if demand keeps increasing, because frequency deviation is independent of supply-demand imbalance. \\
B. Renewable output will necessarily rebound immediately, so electricity price and frequency deviation should both drop to their minimum values. \\
C. Battery power is more likely to switch to strong charging, because a high-load phase implies surplus electricity in the system. \\
D. Thermal generation will become unnecessary, since declining renewable output reduces the need for dispatchable resources. \\
E. Electricity price and frequency deviation may further increase unless additional dispatchable generation or storage discharge is available.

\textbf{Output:} Please output the correct candidate answer.

\textbf{Answer:} \textcolor{answergreen}{\textit{E. The final interval shows persistently high demand
  and continuously weakening renewable support. If this trend continues without additional controllable
  supply, electricity price and frequency deviation are likely to increase further.}}
  
\end{tcolorbox}
\end{figure}

\begin{figure}[h]
\begin{tcolorbox}[
    title=\textbf{Example of the Acting task},
    colback=mycyan,
    colframe=mycyan1,
    boxrule=0.6pt,
    arc=2pt,
    left=5pt,
    right=5pt,
    top=5pt,
    bottom=5pt
]
\textbf{Instruction:} You are an expert in time series analysis and temporal reasoning. A temporal token representing the input time series is provided to the model. Please answer the multiple-choice question based on the temporal evidence, context, and candidate answers. Only one answer is correct.

\textbf{Variable Information:} 
$D$: Load Demand; 
$G$: Thermal Generation; 
$R$: Renewable Output; 
$B$: Battery Power; 
$P$: Electricity Price; 
$F$: Frequency Deviation.

\vspace{2mm}
\begin{center}
    \includegraphics[width=0.92\linewidth]{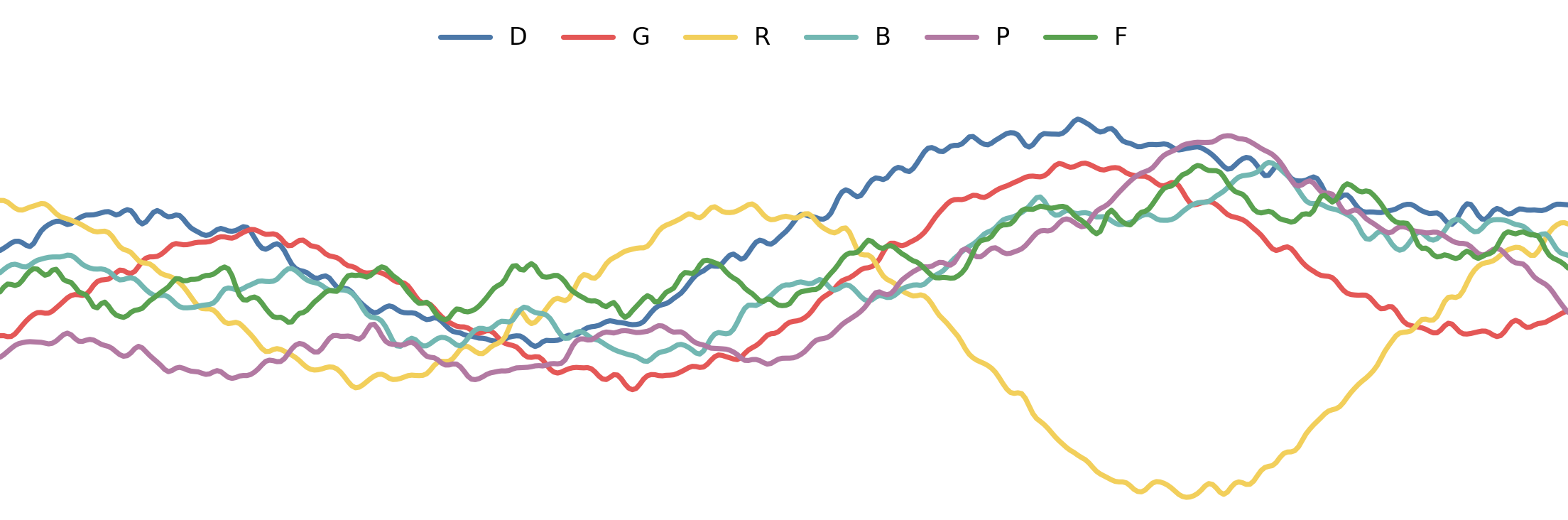}
\end{center}
\vspace{1mm}

\textbf{Context:} 
The data describe the operation of a regional power system. 
The operator monitors load demand, thermal generation, renewable output, battery behavior, electricity price, and frequency deviation to evaluate supply-demand balance and dispatch decisions. 
Battery power reflects storage operation, while electricity price and frequency deviation may indicate system stress. 
The actual temporal state should be inferred from the provided time series representation.

\textbf{Question:} 
Which operational decision best matches the observed system state and dispatch objective?

\textbf{Candidate Answers:} \\
A. Increase battery charging and reduce dispatchable generation, because the system appears to have large surplus supply during the high-load phase. \\
B. Maintain the current dispatch without intervention, because electricity price and frequency deviation provide no useful signal about operating pressure. \\
C. Discharge the battery and increase dispatchable generation to relieve the high-load pressure while monitoring frequency stability. \\
D. Shut down renewable generation completely, because removing renewable output is the most direct way to stabilize price and frequency. \\
E. Reduce both thermal generation and battery output, because lower controllable supply will automatically reduce demand-side pressure.

\textbf{Output:} Please output the correct candidate answer.

\textbf{Answer:} \textcolor{answergreen}{\textit{C. High demand and weak renewable output coincide with
  elevated electricity price and frequency deviation. Increasing dispatchable generation and battery
  discharge is therefore the appropriate response to relieve system pressure.}}
  
\end{tcolorbox}
\end{figure}

\begin{figure}[h]
\begin{tcolorbox}[
    title=\textbf{Failure Case},
    colback=mycyan3,
    colframe=mycyan4,
    boxrule=0.6pt,
    arc=2pt,
    left=5pt,
    right=5pt,
    top=5pt,
    bottom=5pt
]

\textbf{Instruction:} You are an expert in time series analysis and temporal reasoning. A temporal token representing the input time series is provided to the model. Please answer the multiple-choice question based on the temporal evidence, context, and candidate answers. Only one answer is correct.

\textbf{Variable Information:} 
$T$: Surface Temperature; 
$H$: Relative Humidity; 
$P$: Precipitation; 
$W$: Wind Speed; 
$S$: Solar Radiation.

\vspace{2mm}
\begin{center}
    \includegraphics[width=0.92\linewidth]{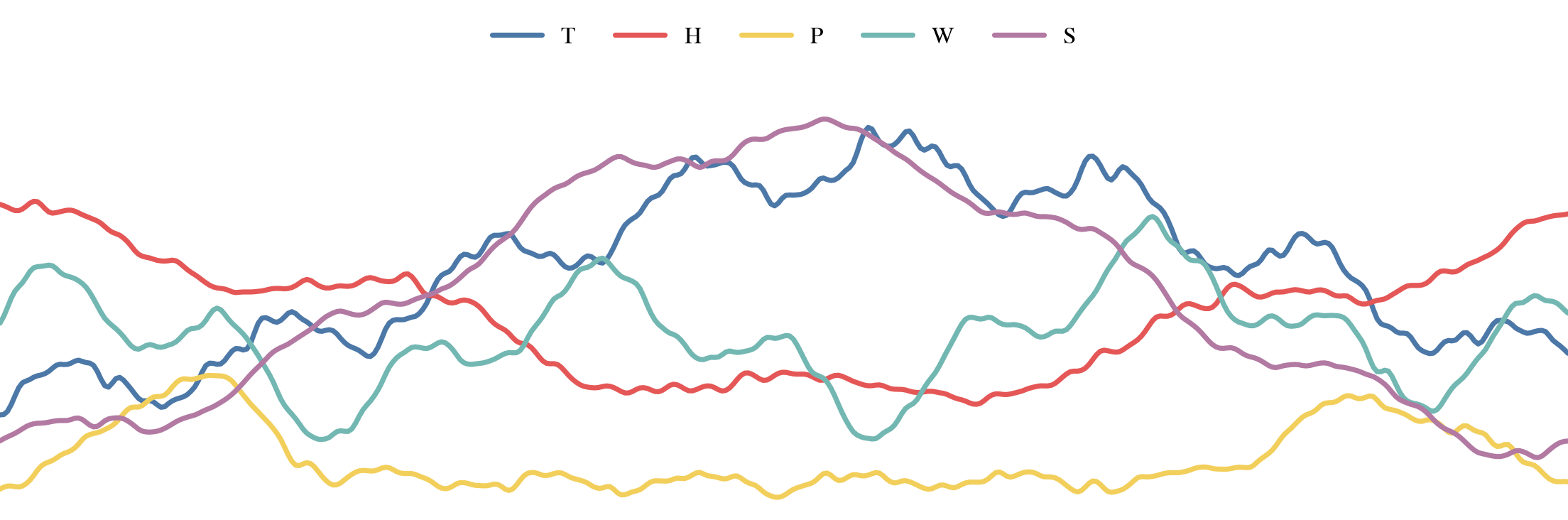}
\end{center}
\vspace{1mm}

\textbf{Context:}
The data describe the operation of a regional weather monitoring system. 
The operator monitors surface temperature, relative humidity, precipitation, wind speed, and solar radiation to evaluate atmospheric conditions and weather evolution. 
Temperature and solar radiation reflect thermal conditions, while humidity and precipitation characterize moisture variation. 
The actual temporal state should be inferred from the provided time series representation.

\textbf{Question:}
Which statement best describes the temporal relationship between surface temperature and solar radiation over the observed period?

\textbf{Candidate Answers:} \\
A. Surface temperature gradually increases throughout the period, while solar radiation decreases after an early peak, indicating a delayed thermal response to weakening radiation. \\
B. Solar radiation increases toward the middle of the period and then decreases, while surface temperature follows a similar pattern with a warming phase followed by cooling. \\
C. Surface temperature decreases during the period, while solar radiation reaches its maximum value near the end, suggesting reduced thermal input and delayed warming. \\
D. Solar radiation remains relatively stable with small fluctuations, while surface temperature exhibits strong variations unrelated to radiation changes. \\
E. Surface temperature and solar radiation both increase continuously throughout the period, showing a consistent upward relationship between thermal conditions and radiation intensity.

\textbf{Answer:} \textcolor{red}{\textit{E. Surface temperature and solar radiation exhibit a broadly aligned upward trend despite several local
  fluctuations.}}

\vspace{1mm}

\textbf{Ground Truth:}
  \textcolor{answergreen}{\textit{B.}}
  
\textbf{Failure Analysis:}
\textit{The model captures the initial increasing trend but fails to recognize the subsequent reversal shared
  by
  surface temperature and solar radiation. It therefore overemphasizes the early local pattern and
  incorrectly predicts a continuous increase. This failure indicates difficulty in tracking coordinated
  phase transitions across variables over the full temporal horizon.}

\end{tcolorbox}
\end{figure}
\end{document}